%% file: arxiv_draft.tex
\pdfoutput=1
\documentclass[acmsmall, screen,nonacm]{acmart}

\makeatletter                   
\def\mdseries@tt{m}             
\makeatother                    

\setcopyright{none}
\acmJournal{PACMPL}
\acmYear{2020}
\copyrightyear{2020}
\acmVolume{4}
\acmNumber{OOPSLA}
\acmArticle{}
\acmMonth{11}
\acmPrice{}
\acmDOI{10.1145/3428230}

\usepackage{booktabs}   %
\usepackage{subcaption} %
\usepackage{natbib}
\usepackage[utf8]{inputenc} %
\usepackage[T1]{fontenc}    %
\usepackage{url}            %
\usepackage{booktabs}       %
\usepackage{amsfonts}       %
\usepackage{nicefrac}       %
\usepackage{microtype}      %
\usepackage{lipsum}
\usepackage{multirow}
\usepackage{makecell}
\usepackage{listings}
\usepackage{cleveref}
\usepackage{paralist}
\usepackage{subcaption}
\usepackage[frozencache]{minted}
\usepackage{pgfplots}
\usepackage{tikz}
\usetikzlibrary{decorations.markings}
\usetikzlibrary{matrix,backgrounds,arrows}
\usepackage{enumitem}
\usepackage{wasysym} 
\usepackage[xcolor,leftbars]{changebar}
\hypersetup{
    breaklinks=true,            
}
\usepackage{breakurl}           

\newcommand{\sname}[1]{{\small \textsc{#1}}}
\newcommand{\scode}[1]{{\small \texttt{#1}}}
\definecolor{mygreen}{HTML}{0F9D58}
\newcommand\boldgreen[1]{{\textcolor{mygreen}{\textbf{#1}}}}
\newcommand\boldmaroon[1]{{\textcolor{Maroon}{\textbf{#1}}}}

\newcommand{\varname}{\sname{VarName}}
\newcommand{\deadcode}{\sname{DeadCode}}
\newcommand{\method}{\sname{DAMP}}
\newcommand{\fullmethod}{Discrete Adversarial Manipulation of Programs}
\newcommand{\norm}[1]{{\left\lVert#1\right\rVert}_{2}}
\newcommand{\slotbox}[1]{\setlength{\fboxrule}{1pt}\fcolorbox{blue}{blue!5}{#1}}

\newcommand\para[1]{\vspace{3pt}\noindent \textbf{#1}}

\begin{document}
\sloppy                         

\title[]{Adversarial Examples for Models of Code}         %

\author{Noam Yefet}
\affiliation{
\institution{Technion}
\country{Israel}
}
\email{snyefet@cs.technion.ac.il}

\author{Uri Alon}
\affiliation{
\institution{Technion}
\country{Israel}
}
\email{urialon@cs.technion.ac.il}

\author{Eran Yahav}
\affiliation{
\institution{Technion}
\country{Israel}
}
\email{yahave@cs.technion.ac.il}

\input{abstract}



\maketitle

\input{intro.tex}

\input{overview}

\input{background.tex}

\input{adversarial}

\input{defense.tex}

\input{eval.tex}

\input{eval_examples}

\input{related}

\input{conclusion}
\input{ack}

\appendix
\section{Appendix - Additional Examples}
\label{app:examples}

\input{example_escape}

\bigbreak \bigbreak \bigbreak \bigbreak
\input{example_contains}

\input{example_count}

\bigbreak \bigbreak \bigbreak \bigbreak
\input{example_propertyEmitter.tex}

\input{example_proxyGenerator.tex}

\input{example_gnn_SourceType2.tex}

\input{example_gnn_enum_to_enum.tex}

\clearpage
\bibliography{bib}

\end{document}

%% file: abstract.tex
\begin{abstract}
Neural models of code have shown impressive results when performing tasks such as predicting method names and identifying certain kinds of bugs.
We show that these models are vulnerable to \emph{adversarial examples}, and introduce a novel approach for \emph{attacking} trained models of code using adversarial examples.
The main idea of our approach is to force a given trained model to make an incorrect prediction, as specified by the adversary, by introducing small perturbations that do not change the program's semantics, thereby creating an adversarial example.
To find such perturbations, we present a new technique for \fullmethod{} (\method{}). {\method} works by deriving the desired prediction with respect to the model's \emph{inputs}, while holding the model weights constant, and following the gradients to slightly modify the input code.

We show that our \method{} attack is effective across three neural architectures: \sname{code2vec}, \sname{GGNN}, and \sname{GNN-FiLM}, in both Java and C\#. 
Our evaluations demonstrate that \method{} has up to $89\%$ success rate in changing a prediction to the adversary's choice (a targeted attack) and a success rate of up to~$94\%$ in changing a given prediction to any incorrect prediction (a non-targeted attack). 
To defend a model against such attacks, we empirically examine a variety of possible defenses and discuss their trade-offs.
We show that some of these defenses can dramatically drop the success rate of the attacker, with a minor penalty of $2\%$ relative degradation in accuracy when they are not performing under attack.

Our code, data, and trained models are available at \url{https://github.com/tech-srl/adversarial-examples} .
\end{abstract}

%% file: intro.tex
\input{intro_figure.tex}

\section{Introduction}

Neural models of code have achieved state-of-the-art performance on various tasks such as the prediction of variable names and types~\cite{jsnice2015, pigeon, phog16, allamanis2018learning}, code summarization~\cite{conv16, alon2019code2seq, fernandes2018structured}, code generation~\cite{brockschmidt2018generative, murali2018bayou, alon2019structural}, code search~\cite{sachdev2018retrieval, liu201search, cambronero2019deep}, and bug finding~\cite{pradel2018deepbugs, rice2017detecting, scott2019getafix}.

In other domains such as computer vision, deep models have been shown to be vulnerable to \emph{adversarial examples}~\cite{szegedy2013intriguing, goodfellow2014explaining}. Adversarial examples are inputs crafted by an adversary to force a trained neural model to make a certain (incorrect) prediction. 
The generation of adversarial examples was demonstrated for image classification~\cite{szegedy2013intriguing, goodfellow2014explaining, goodfellow2014generative, nguyen2015deep,mirza2014conditional, papernot2016limitations, kurakin2016adversarial, papernot2017practical, moosavi2016deepfool} and for other domains~\cite{carlini2018audio, taori2019targeted, ebrahimi2017hotflip, alzantot2018generating, alzantot2018did, belinkov2017synthetic,pruthi2019combating}. The basic idea underlying many of the techniques is to add specially-crafted noise to a correctly labeled input, such that the model under attack yields a desired incorrect label when presented with the modified input (i.e., with the addition of noise). 
Adding noise to a \emph{continuous object} to change the prediction of a model is relatively easy to achieve mathematically. For example, for an image, this can be achieved by changing the intensity of pixel values~\cite{szegedy2013intriguing, goodfellow2014explaining}. Unfortunately, this does not carry over to the domain of programs, since a program is a \emph{discrete object} that must maintain semantic properties.

In this paper, we present a novel approach for generating adversarial examples for neural models of code. 
More formally:

\subsection{Goal} Given a program $\mathcal{P}$ and a correct prediction $y$ made by a model $\mathcal{M}$, such that: $\mathcal{M}\left(\mathcal{P}\right)=y$, our goal is to find a semantically equivalent program $\mathcal{P}'$ such that $\mathcal{M}$ makes a given adversarial prediction $y_{bad}$ of the adversary's choice: $\mathcal{M}\left(\mathcal{P'}\right)=y_{bad}$.

The main challenge in tackling the above goal lies in exploring the vast space of programs that are semantically equivalent to $\mathcal{P}$, and finding a program for which $\mathcal{M}$ will predict $y_{bad}$.

Generally, we can define a set of semantic-preserving transformations, which in turn induce a space of semantically equivalent programs. For example, we can: 
\begin{inparaenum}[(i)]
	\item rename variables and
	\item add dead code.
\end{inparaenum}
There are clearly many other semantic preserving transformations (e.g., re-ordering independent statements), but their application would require a deeper analysis of the program to guarantee that they are indeed semantic preserving. 
In this paper, therefore, we focus on the above two semantic-preserving transformations, which can be safely applied without any semantic analysis. 

One na\"ive approach for exploring the space of equivalent programs is to randomly apply transformations using brute-force. We can apply transformations randomly to generate new programs and use the model to make a prediction for each generated program. %
However, the program space to be explored is exponentially large, making exhaustive exploration prohibitively expensive.

\input{onehot_tikz_intro}
\subsection{Our Approach} We present a new technique called \fullmethod{} (\method{}). 
The main idea in \method{} is to select semantic preserving perturbations by deriving the output distribution of the model with respect to the model's input and following the gradient to modify the input, while keeping the model weights constant. 
Given a desired adversarial label $y_{bad}$ and an existing variable name, we derive the loss of the model with $y_{bad}$ as the correct label, with respect to the one-hot vector of the input variable. We then take the argmax of the resulting gradient to select an alternative variable name, rename the original variable to the alternative name, check whether this modification changes the output label to the desired adversarial label, and continue iteratively.
This process is illustrated in \Cref{fig:onehot-intro}, and detailed in \Cref{sec:approach}.

This iterative process allows \method{} to modify the program in a way that preserves its semantics but will cause a model to make adversarial predictions.
 We show that models of code are susceptible to \emph{targeted attacks} that force a model to make a specific incorrect prediction chosen by the adversary, as well as to simpler \emph{non-targeted attacks} that force a model to make \emph{any incorrect prediction} without a specific target prediction in mind. Our approach is a ``white-box'' approach, since it assumes the attacker has access to either the model under attack or to a similar model. 
 \footnote{As recently shown by \citet{wallace2020stealing}, this is a reasonable assumption. An attacker can imitate the model under attack by: training an imitation model using labels achieved by querying the original model; crafting adversarial examples using the imitation model; and transferring these adversarial examples back to the original model.} 
 Under this assumption, our approach is general and \emph{is applicable to any model that can be derived with respect to its inputs} i.e., any neural model. We do not make any assumptions about the internal details or specific architecture of the model under attack.  

To mitigate these attacks, we evaluate and compare a variety of \emph{defensive} approaches. Some of these defenses work by re-training the model using another loss function or a modified version of the same dataset. Other defensive approaches are ``modular'', in the sense that they can be placed in front of an already-trained model, identify perturbations in the input, and feed a masked version of the input into the vulnerable model. 
These defense mechanisms allow us to trade off the accuracy of the original model for improved robustness. 

\para{Main Contributions} The contributions of this paper are:
\begin{itemize}
	\item The first technique for generating \emph{targeted} adversarial examples for models of code. Our technique, called \fullmethod{} (\method{}), is general and only requires that the attacker is able to compute gradients in the model under attack (or in a similar model). {\method{}} is effective in generating both targeted and non-targeted attacks. 
	\item An experimental evaluation of attacks on three neural architectures: code2vec~\cite{alon2019code2vec}, \sname{\sname{GGNN}}~\cite{allamanis2018learning}, and \sname{GNN-FiLM}~\cite{brockschmidt2018generative} in two languages: Java and C\#. Our evaluation shows that our adversarial technique can change a prediction according to the adversary's will (``targeted attack'') up to $89\%$ of the time, and is successful in changing a given prediction to an incorrect prediction (``non-targeted attack'') $94\%$ of the time. 
	\item A thorough evaluation of techniques for defending models of code against attacks that perturb names, and an analysis of their trade-offs. When some of these defenses are used, the success rate of the attack drops drastically for both targeted and non-targeted attacks, with a minor penalty of $2\%$ in accuracy.

\end{itemize}

%% file: intro_figure.tex
\begin{figure*}[t]
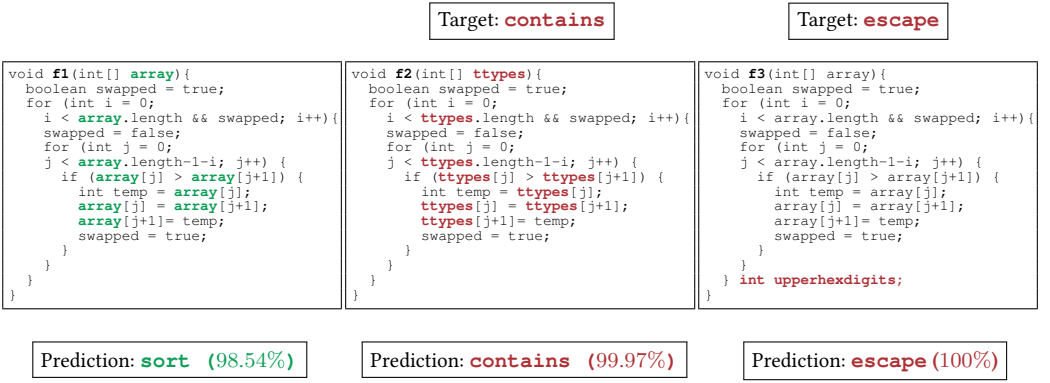

\begin{minipage}{\textwidth}
\centering

\begin{tabular}{lll}
\textbf{Correctly predicted example}

&

\multicolumn{2}{c}{
\textbf{Adversarial perturbations}
}

\\ \\

&
\hspace{6mm}
\fbox{\footnotesize Target: \boldmaroon{\texttt{contains}}} 
&
\hspace{8mm}
\fbox{\footnotesize Target: \boldmaroon{\texttt{escape}}} \\

\hspace{-2mm}
\begin{subfigure}[t]{0.325\textwidth}
\begin{minted}[fontsize=\fontsize{5.5}{5.5}, frame=single,framesep=2pt,escapeinside=||]{text}
void |\textbf{f1}|(int[] |\boldgreen{array}|){
  boolean swapped = true;
  for (int i = 0; 
    i < |\boldgreen{array}|.length && swapped; i++){
    swapped = false;
    for (int j = 0; 
    j < |\boldgreen{array}|.length-1-i; j++) {
      if (|\boldgreen{array}|[j] > |\boldgreen{array}|[j+1]) {
        int temp = |\boldgreen{array}|[j];
        |\boldgreen{array}|[j] = |\boldgreen{array}|[j+1];
        |\boldgreen{array}|[j+1]= temp;
        swapped = true;
      }
    }
  }
}
\end{minted}
\end{subfigure}
&
\hspace{-5.2mm}

\begin{subfigure}[t]{0.335\textwidth}
\begin{minted}[fontsize=\fontsize{5.5}{5.5}, frame=single,framesep=2pt,escapeinside=||]{text}
void |\textbf{f2}|(int[] |\boldmaroon{ttypes}|){
  boolean swapped = true;
  for (int i = 0; 
    i < |\boldmaroon{ttypes}|.length && swapped; i++){
    swapped = false;
    for (int j = 0; 
    j < |\boldmaroon{ttypes}|.length-1-i; j++) {
      if (|\boldmaroon{ttypes}|[j] > |\boldmaroon{ttypes}|[j+1]) {
        int temp = |\boldmaroon{ttypes}|[j];
        |\boldmaroon{ttypes}|[j] = |\boldmaroon{ttypes}|[j+1];
        |\boldmaroon{ttypes}|[j+1]= temp;
        swapped = true;
      }
    }
  }
}
\end{minted}
\end{subfigure}
&
\hspace{-4mm}

\begin{subfigure}[t]{0.325\textwidth}
\begin{minted}[fontsize=\fontsize{5.5}{5.5}, frame=single,framesep=2pt,escapeinside=||]{text}
void |\textbf{f3}|(int[] array){
  boolean swapped = true; 
  for (int i = 0; 
    i < array.length && swapped; i++){
    swapped = false;
    for (int j = 0; 
    j < array.length-1-i; j++) {
      if (array[j] > array[j+1]) {
        int temp = array[j];
        array[j] = array[j+1];
        array[j+1]= temp;
        swapped = true;
      }
    }
  } |\boldmaroon{int upperhexdigits;}|
}
\end{minted}
\end{subfigure}
\\
\\
\hspace{2mm}
\fbox{\footnotesize Prediction: \boldgreen{\texttt{sort ($98.54\%$)}}} 
&
\hspace{-3mm}
\fbox{\footnotesize Prediction: \boldmaroon{\texttt{contains ($99.97\%$)}}} 

&
\hspace{2mm}
\fbox{\footnotesize Prediction: \boldmaroon{\texttt{escape} ($100\%$)}} \\

\end{tabular}
\end{minipage}
\caption{A Java snippet \scode{f1} is classified correctly as \scode{sort} by the model of \href{https://code2vec.org}{\url{code2vec.org}}. Given \scode{f1} and the target \scode{contains}, our approach generates \scode{f2} by renaming \scode{array} to \scode{ttypes}. Given the target \scode{escape}, our approach generates \scode{f3} by adding an unused variable declaration of \scode{int upperhexdigits}. Additional examples can be found in \Cref{app:examples}.}
\label{IndexOfOriginal}
\end{figure*}

%% file: onehot_tikz_intro.tex
\begin{figure}
\footnotesize
\input{onehot_tabular.tex}

\caption{Perturbing a variable name: the original variable name is represented as a one-hot vector over the variable-name vocabulary. After perturbation, the vector is no longer one-hot. We apply argmax to find the most likely adversarial name, resulting with another one-hot vector over the variable-name vocabulary.}
\label{fig:onehot-intro}
\end{figure}

%% file: onehot_tabular.tex
\begin{tabular}{l}
	\begin{tikzpicture}[
array/.style={matrix of nodes,nodes={draw, minimum size=5mm, fill=white},column sep=-\pgflinewidth, row sep=0.5mm, nodes in empty cells, 
},>=triangle 45]

\matrix[array] (array) {
  &   &   &   &   &   &   &   &   &  \\
  };
\node[draw, fill=blue, minimum size=5mm] at (array-1-6) (box) {};

\node [align=center, anchor=south] at ([yshift=5mm]array-1-9.north west) (4) {{\footnotesize Original variable name}};
\draw (4)--(box);

\draw[->] (0,-0.6) -- (0,-1.2) node[black,midway,right] {$v' = \bar{v} - \eta \cdot \nabla_{\bar{v}} J(\theta, x, y_{bad})
$};%

\matrix[array] (array) {
  &   &   &   &   &   &   &   &   &  \\
  };
  
\node[draw, fill=blue, minimum size=5mm] at (array-1-6) (box) {};
  
\end{tikzpicture}
\\ 
\begin{tikzpicture}[font=\ttfamily,
array/.style={matrix of nodes,nodes={draw, minimum size=5mm, fill=white},column sep=-\pgflinewidth, row sep=0.5mm, nodes in empty cells, 
},>=triangle 45]
\matrix[array] (array) {
  &   &   &   &   &   &   &   &   &  \\
  };
\draw[->] (0,-0.6) -- (0,-1.2) node[black,midway,right] {$argmax$};

\matrix[array] (array) {
  &   &   &   &   &   &   &   &   &  \\
  };

\node[draw, fill=blue!20, minimum size=5mm] at (array-1-2) (box) {};
\node[draw, fill=blue!70, minimum size=5mm] at (array-1-3) (box) {};  
\node[draw, fill=blue!30, minimum size=5mm] at (array-1-4) (box) {};
\node[draw, fill=blue!40, minimum size=5mm] at (array-1-5) (box) {};
\node[draw, fill=blue!30, minimum size=5mm] at (array-1-8) (box) {};
\node[draw, fill=blue!50, minimum size=5mm] at (array-1-9) (box) {};
\node[draw, fill=blue!20, minimum size=5mm] at (array-1-6) (box) {};
\end{tikzpicture}
\\
\begin{tikzpicture}[
array/.style={matrix of nodes,nodes={draw, minimum size=5mm, fill=white},column sep=-\pgflinewidth, row sep=0.5mm, nodes in empty cells, 
},>=triangle 45]
\matrix[array] (array) {
  &   &   &   &   &   &   &   &   &  \\
  };
\node[draw, fill=blue, minimum size=5mm] at (array-1-6) (box) {};
\matrix[array] (array) {
  &   &   &   &   &   &   &   &   &  \\
  };
   
\node[draw, fill=blue, minimum size=5mm] at (array-1-3) (box) {};  
\node [align=center, anchor=north] at ([yshift=-5mm]array-1-6.south west) (4) {{\footnotesize Adversarial variable name}};
\draw (4)--(box);v
\end{tikzpicture}
\end{tabular}

%% file: overview.tex
\input{gnn_intro_figure.tex}

\section{Overview}
In this section, we provide an informal overview.

\subsection{Motivating Examples}
We begin by demonstrating our technique on two examples, which address two different tasks, using two different neural models, and in two programming languages (Java and C\#).

\para{Bypass Semantic Labeling (\sname{code2vec} - Java)}
We demonstrate how our approach can force the \sname{code2vec} \cite{alon2019code2vec} model to predict a label of our choice. Consider the code snippet \scode{f1} of~\Cref{IndexOfOriginal}. This code snippet sorts a given array. The \sname{code2vec} model \cite{alon2019code2vec} applied to this code snippet predicts the correct name, \scode{sort}, with a probability of $98.54\%$. 

\begin{sloppypar}
	Given the code snippet \scode{f1} and the arbitrarily chosen adversarial target \scode{contains}, our approach finds that renaming the original variable \scode{array} to \scode{ttypes} in \scode{f2} forces the model to predict the label \scode{contains} with an even higher probability, although the functionality remained the same. We denote this as a \varname{} attack.
\end{sloppypar}

Given the code snippet \scode{f1} and the adversarial target \scode{escape}, our approach finds that adding an unused variable (i.e., dead code) named \scode{upperhexdigits} to the end of \scode{f1} forces the model to predict the label \scode{escape} with a probability of $100\%$ (see \scode{f3} in \Cref{IndexOfOriginal}).
We denote this attack as a \deadcode{} targeted attack.

In general, bypassing semantic labeling models can allow a variety of malicious behaviors. For example, a malicious developer can spread malicious code to users by having the code classified as ``benign'' by a malware detection model \cite{david2019neural, saxe2015deep, wang2017adversary, arp2014drebin, grosse2016adversarial}. In this paper, we demonstrate the ability to bypass semantic labeling by applying our attack on the \sname{code2vec} model (\Cref{IndexOfOriginal}), forcing the model to predict a label of our choice.

\para{Bypass Bug Detection (\sname{VarMisuse} - C\#)}
As another example, we demonstrate how our approach can force a Graph Neural Network (GNN) model to choose an obvious bug as the correct completion.
In \Cref{gnnDestinationTypeCorrect}, a GNN model trained on the \sname{VarMisuse} task \cite{allamanis2018learning, brockschmidt2019gnn} in C\# correctly chooses to ``fill the blank'' using the field \scode{DestinationType} inside the method \scode{Equals}. By renaming a local variable called \scode{destination} in \emph{another method} to the specific name \scode{scsqbhj} (\Cref{gnnDestinationTypeIncorrect}), the model chooses the incorrect field \scode{SourceType} in the method \scode{Equals}. The fields \scode{DestinationType} (correct) and \scode{SourceType} (incorrect) both have the same type; thus, the code still compiles and the attack causes a real bug in \scode{Equals}. 

More generally, bypassing a bug detection model \cite{pradel2018deepbugs, rice2017detecting, scott2019getafix} can allow a malicious developer inside an organization or inside an open-source project to intentionally introduce bugs. In this paper, we demonstrate this ability using the \sname{VarMisuse} on Graph Neural Networks (GNNs) (\Cref{gnnDestinationType}), forcing the model to choose an incorrect (but type-correct) variable.

In addition to the \sname{code2vec} and \sname{VarMisuse} tasks that we address in this paper, 
we believe adversarial examples can be applied to neural code search \cite{sachdev2018retrieval, liu201search, cambronero2019deep}. A developer can attract users to a specific library or an open-source project by introducing code that will be disproportionately highly ranked by a neural code search model.

\subsection{{\fullmethod{} (\method)}}
\label{subsec:method}

Consider the code snippet \scode{f1} of~\Cref{IndexOfOriginal} that sorts a given array. The \sname{code2vec} model \cite{alon2019code2vec} applied to this code snippet predicts the correct name, \scode{sort}. 
Our goal is to find \emph{semantically equivalent} snippets that will cause an underlying model to yield an incorrect target prediction of our choice.

\para{Gradient-Based Exploration of the Program Space} We need a way to guide exploration of the program space towards a specific desired target label (in a targeted attack), or away from the original label (in a non-targeted attack). 

In standard stochastic gradient descent (SGD)-based training of neural networks, the weights of the network are updated to minimize the loss function. The gradient is used to guide the update of the network weights to minimize the loss. However, what we are trying to determine is not an update of the network's weights, but rather an ``update'' of the network's \emph{inputs}. A natural way to obtain such guidance is to derive the desired prediction with respect to the model's \emph{inputs} while holding the model weights constant and follow the gradient to modify the inputs.  

In settings where the input is continuous (e.g., images), modifying the input can be done directly by adding a small noise value and following the direction of the gradient towards the desired target label (targeted), or away from the original label (non-targeted). A common technique used for images is the \emph{Fast Gradient Signed Method} (FGSM) \cite{goodfellow2014explaining} approach, which modifies the input using a small fixed $\epsilon$ value.

\para{Deriving with Respect to a Discrete Input} In settings where the input is discrete, the first layer of a neural network is typically an embedding layer that embeds discrete objects, such as names and tokens, into a continuous space \cite{alon2019code2seq, conv16, codenn16}. The input is the index of the symbol, which is used to look up its embedding in the embedding matrix.
The question for discrete inputs is therefore: \emph{what does it mean to derive with respect to the model's inputs?} 

One approach is to %
derive with respect to the \emph{embedding vector}, which is the result of the embedding layer. In this approach, after the gradient is obtained, we need to reflect the update of the embedding vector back to discrete-input space. This can be done by looking for the nearest-neighbors of the updated embedding vector in the original embedding space, and finding a nearby vector that has a corresponding discrete input. In this approach, there is no guarantee that following the gradient is the best step. 

In contrast, our \fullmethod{} (\method{}) approach derives with respect to a one-hot vector that represents the \emph{distribution} over discrete values (e.g., over variable names). 
Instead of deriving by the input itself, the gradient is taken with respect to the \emph{distribution} over the inputs.
Intuitively, this allows us to directly obtain the best discrete value for following the gradient.%

\para{Targeted Gradient-based Attack}
Using our gradient-based method, we explore the space of semantically equivalent programs \emph{directly toward} a desired adversarial target. For example, given the code snippet \scode{f1} of~\Cref{IndexOfOriginal} and the desired target label \scode{contains}, our approach for generating adversarial examples automatically infers the snippet \scode{f2} of~\Cref{IndexOfOriginal}. Similarly, given the target label \scode{escape}, our approach automatically infers the snippet \scode{f3} of~\Cref{IndexOfOriginal}.

All code snippets of~\Cref{IndexOfOriginal} are semantically equivalent. The only difference between \scode{f1} and \scode{f2} is the name of the variables. Specifically, these snippets differ only in the name of a single variable, which is named \scode{array} in \scode{f1} and \scode{ttypes} in \scode{f2}. Nevertheless, when \scode{array} is renamed to \scode{ttypes}, the prediction made by \sname{code2vec} changes to the desired (adversarial) target label \scode{contains}. 
The difference between \scode{f1} and \scode{f3} is the addition of a single variable declaration \scode{int upperhexdigits}, which is never used in the code snippet. Nevertheless, adding this declaration changes the prediction made by the model to the desired (adversarial) target label \scode{escape}.

%% file: gnn_intro_figure.tex
\begin{figure*}[t]
\begin{minipage}{\textwidth}
\centering

\begin{tabular}{cc}
\textbf{Correctly predicted example}

\hspace{3mm}
&

\textbf{Adversarial perturbation} \\
\\
& \fbox{\footnotesize Target: \boldmaroon{\texttt{SourceType}}} \\
\hspace{-2mm}

\begin{subfigure}[t]{0.49\textwidth}
\begin{minted}[fontsize=\fontsize{8}{8}, frame=single,framesep=2pt,escapeinside=||]{text}
struct TypePair : IEquatable<TypePair>
{
  public static TypePair Create<TSource, 
      TDestination>(TSource source, 
        TDestination |\boldgreen{destination}|, ...)
  {
    ...
  }
  
  ...
  
  public Type SourceType { get; }
  public Type DestinationType { get; }
  public bool |\textbf{Equals}|(TypePair other) =>
    SourceType == other.SourceType 
    && DestinationType 
        == other.|\slotbox{\boldgreen{DestinationType}}|;
}
\end{minted}
\caption{}
\label{gnnDestinationTypeCorrect}
\end{subfigure}
&
\hspace{-2mm}
\begin{subfigure}[t]{0.49\textwidth}
\begin{minted}[fontsize=\fontsize{8}{8}, frame=single,framesep=2pt,escapeinside=||]{text}
struct TypePair : IEquatable<TypePair>
{
  public static TypePair Create<TSource, 
      TDestination>(TSource source, 
        TDestination |\boldmaroon{scsqbhj}|, ...)
  {
    ...
  }
  
  ...
  
  public Type SourceType { get; }
  public Type DestinationType { get; }
  public bool |\textbf{Equals}|(TypePair other) => 
    SourceType == other.SourceType 
    && DestinationType 
        == other.|\slotbox{\boldmaroon{SourceType}}|;
}
\end{minted}
\caption{}
\label{gnnDestinationTypeIncorrect}
\end{subfigure}

\end{tabular}
\end{minipage}
\caption{A C\# \sname{VarMisuse} example which is classified correctly as \scode{DestinationType} in the method \scode{Equals} by the \sname{GGNN} model of \citet{allamanis2018learning}. Given the code in \Cref{gnnDestinationTypeCorrect} and the target \scode{SourceType}, our approach renames a local variable \scode{destination} in \emph{another method} to the specific name \scode{scsqbhj}, making the model predict the wrong variable in the method \scode{Equals}, thus (``maliciously'') introducing a real bug in the method \scode{Equals}. Additional examples are shown in \Cref{app:examples}.}
\label{gnnDestinationType}
\end{figure*}

%% file: background.tex
\section{Background} 

In this section we provide fundamental background on neural networks and adversarial examples. \\

\subsection{Training Neural Networks}
\label{subsec:background_training}
A Neural Network (NN) model can be viewed as a function $f_{\theta} : X \rightarrow{} Y$, where $X$ is the input domain (image, text, code, etc.) and $Y$ is usually a finite set of labels. Assuming a perfect classifier $h^*: X \rightarrow{} Y$, the goal of the function $f_{\theta}$ is to assign the correct label $y \in Y$ (which is determined by $h^*$) for each input $x \in X$. In order to accomplish that, $f_{\theta}$ contains a set of \emph{trainable weights}, denoted by $\theta$, which can be adjusted to fit a given labeled training set 
$T = \{ (x,y) | x \in \bar{X} \subset X, y \in Y, y = h^*(x) \} $. 
The process of adjusting $\theta$ (i.e., \emph{training}) is done by solving an optimization problem defined by a certain loss function $ J(\theta, x, y)$; this is usually mean squared error (MSE) or cross entropy and is used to estimate the model's generalization ability:
\begin{equation}
    \theta^* = \operatorname*{argmin}_\theta \sum_{(x,y) \in T} J(\theta, x, y)
\end{equation}
One of the most common algorithms to approximate the above problem is \emph{gradient descent} \cite{cauchy1847methode} using \emph{backpropagation} \cite{kelley1960gradient}. When gradient descent is used for training, the following update rule is applied repeatedly to update the model's weights:
\begin{equation}
    \theta_{t+1} = \theta_{t} - \eta \cdot \nabla_\theta J(\theta_{t}, x, y)
    \label{eq:gd}
\end{equation}
where $\eta \in \mathcal{R}$ is a small scalar hyperparameter called \emph{learning rate}, for example, $0.001$. 
Intuitively, the gradient descent algorithm can be viewed as taking small steps in the direction of the steepest descent, until a (possibly local) minimum is reached. This process is illustrated in \Cref{fig:gradientDescentIllustration}, where $\theta$ contains a single trainable variable $\theta \in \mathcal{R}$. %
\input{sgd_illustration}

\subsection{Adversarial Examples}

Neural network models are very popular and have been
applied in many domains, including computer vision \cite{NIPS2012_4824,simonyan2014very, szegedy2015going, he2016deep}, natural language \cite{mikolov2010recurrent, hochreiter1997long, cho2014learning}, and source code \cite{jsnice2015, phog16, bavishi2018context2name, pigeon, conv16, alon2019code2seq, brockschmidt2018generative, murali2018bayou, lu2017data, sachdev2018retrieval, liu201search, pradel2018deepbugs, rice2017detecting, allamanis2018learning}. 

Although neural networks have shown astonishing results in many domains, they were found to be vulnerable to adversarial examples. An adversarial example is an input which intentionally forces a given trained model to make an incorrect prediction. For neural networks that are trained on continuous objects like images and audio, the adversarial examples are usually achieved by applying a small perturbation on a given input image \cite{szegedy2013intriguing, goodfellow2014explaining, carlini2018audio}. This perturbation is found using gradient-based methods: usually by deriving the desired loss with respect to the neural network's inputs.

In \emph{discrete} problem domains such as natural language or programs, generating adversarial examples is markedly different. Specifically, the existing techniques and metrics do not hold, because discrete symbols cannot be perturbed with imperceptible adversarial noise. 
Additionally, in discrete domains, deriving the loss with respect to its discrete input results in a gradient of zeros.

Recently, attempts were made to find adversarial examples in the domain of Natural Language Processing (NLP). Although adversarial examples on images are easy to generate, the generation of adversarial text is harder. This is due to the discrete nature of text and the difficulty of generating semantic-preserving perturbations. One of the approaches to overcome this problem is to replace a word with a synonym \cite{alzantot2018generating}. Another approach is to insert typos into words by replacing a few characters in the text \cite{belinkov2017synthetic, ebrahimi2017hotflip}.
However, these NLP approaches allow only \emph{non-targeted} attacks.

Additionally, NLP approaches cannot be directly applied to code. 
NLP models don't take into account the unique properties of code, such as: multiple occurrences of variables, semantic and syntactic patterns in code, the relation between different parts of the code, the readability of the entire code, and whether or not the code still compiles after the adversarial mutation. Therefore, applying NLP methods on the code domain makes the hard problem even harder.

%% file: sgd_illustration.tex
\begin{figure}
\begin{tikzpicture}
	\draw [<->,thick] (-0.6,2.25) node (yaxis) [above] {$J$}
        |- (3,0) node (xaxis) [right] {$\theta$};
	  \draw[thick] (1.5,0.5) parabola (3,2.25) ;
	  \draw[thick] (1.5,0.5) parabola (0,2.25) ;
	  \coordinate (v) at (1.5,0.5) ;
	  \coordinate (t1) at (0.16,1.9) ;
	  \coordinate (t1d) at (0.33,1.5) ;
	  \coordinate (t2) at (0.44,1.35) ;
	  \coordinate (t2d) at (0.69,0.9) ;
	  \coordinate (t3) at (1,0.7) ;
	  \coordinate (t3d) at (1.4,0.4) ;
	  \coordinate (a) at (0.42,1.6) ; 
	  \draw[->,thick,red] (t1) -- (t1d) node[black,midway,sloped,left,rotate=65] {{\tiny $ -\nabla_{v} J$}}; 
	  \draw[->,thick,red] (t2) -- (t2d) node[black,midway,sloped,left,rotate=65] {{\tiny $ -\nabla_{v} J$}};
	  \draw[->,thick,red] (t3) -- (t3d) node[black,midway,sloped,below,rotate=35] {{\tiny $ -\nabla_{v} J$}};
	  \fill[blue] (v) circle (2pt) ;
	  \fill[blue] (t1) circle (2pt) ;
	  \fill[blue] (t2) circle (2pt) ;
	  \fill[blue] (t3) circle (2pt) ;
	  \draw[gray,dashed] (v) -- (xaxis -| v) node[black,below] {{\footnotesize$\theta_4$}};
	  \draw[gray,dashed] (t1) -- (xaxis -| t1) node[black,below] {{\footnotesize$\theta_1$}};
	  \draw[gray,dashed] (t2) -- (xaxis -| t2) node[black,below] {{\footnotesize$\theta_2$}};
	  \draw[gray,dashed] (t3) -- (xaxis -| t3) node[black,below] {{\footnotesize$\theta_3$}};
\end{tikzpicture}
\caption{Illustration of gradient descent, subscripts denote different time steps: in each step, the gradient is computed w.r.t. $\theta_t$ for calculating a new $\theta_{t+1}$ by updating $\theta_t$ towards the opposite direction of the gradient, until we reach a (possibly local) minimum value of $J$.}	    
\label{fig:gradientDescentIllustration}
\end{figure}
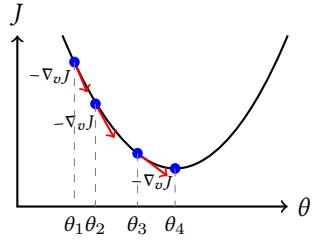

%% file: adversarial.tex
\section{Adversarial Examples for Models of Code}\label{sec:approach}
In this section we describe the process of generating adversarial examples using our \method{} approach. %

\subsection{Definitions}
\label{subsec:definitions}
Suppose we are given a trained model of code. The given model can be described as a function $f_{\theta}: \mathcal{C} \rightarrow \mathcal{Y}$, where $\mathcal{C}$ is the set of all code snippets and $\mathcal{Y}$ is a set of labels.

Given a code snippet $c \in \mathcal{C}$ that the given trained model predicts as $y\in\mathcal{Y}$, i.e., $f_{\theta}\left(c\right)=y$, we denote by $y_{bad}\in\mathcal{Y}$ the \emph{adversarial label} chosen by the attacker. 
Following the definitions of \citet{bielik2020adversarial}:
let $\Delta\left(c\right)$ be a set of valid modifications of the code snippet $c$, and let $\delta\left(c\right)$ be a new input obtained by applying a modification $\delta: \mathcal{C} \rightarrow \mathcal{C}$  to $c$, such that $\delta \in \Delta\left(c\right)$ .
For instance, if $c$ is the code snippet in \Cref{gnnDestinationTypeCorrect}, then \Cref{gnnDestinationTypeIncorrect} shows $\delta_{\scode{destination}\rightarrow\scode{scsqbhj}}\left(c\right)$ -- the same code snippet where $\delta_{\scode{destination}\rightarrow\scode{scsqbhj}}$ is the modification that renames the variable \scode{destination} to \scode{scsqbhj}. As a result, the prediction of the model changes from $y$$=$\scode{DestinationType} to $y_{bad}$$=$\scode{SourceType}, and  $y=f_{\theta}\left(c\right) \neq f_{\theta}\left(\delta_{\scode{destination}\rightarrow\scode{scsqbhj}}\left(c\right)\right)=y_{bad}$.

In this paper, we focus on perturbations that rename local variables or add dead code.
For brevity, in this section we focus only on generating adversarial examples by renaming of the original variables (\varname). 
Thus, we define the possible perturbations that we consider as 
$\Delta\left(c\right) = \{\delta_{v \rightarrow v'} \mid \forall c\in \mathcal{C}: \delta_{v \rightarrow v'}\left(c\right) = c_{v \rightarrow v'} \}$,
where $c_{v \rightarrow v'}$ is the code snippet $c$ in which the variable $v$ was renamed to $v'$.

Our approach can also be applied to a code snippet without changing the existing names, and instead adding a redundant variable declaration (\deadcode{} insertion, see \Cref{subsubsec:strategies}). In such a case, our approach can be similarly applied by choosing an initial random name for the new redundant variable, and selecting this variable as the variable we wish to rename.

We define 
\textbf{$Var(c)$} as the set of all local variables existing in $c$. 
The adversary's objective is to 
 thus select a single variable $v \in Var(c)$ and an alternative name $v'$, such that renaming $v$ to $v '$ will make the model predict the adversarial label. The most challenging step is to find the right alternative name $v'$ such that $f_{\theta}\left(\delta_{v \rightarrow v'}\left(c\right)\right)=y_{bad}$.

\paragraph{Minimality of Perturbation}
In addition to semantic equivalence, we also require that the programs ``before'' and ``after'' the perturbation be as similar as possible, i.e., $c\approx \delta\left(c\right)$. While this constraint is well-defined formally and intuitively in continuous domains such as images and audio \cite{szegedy2013intriguing, carlini2018audio}, in discrete domains (e.g., programs) the existing definitions do not hold. The main reason is that every change is perceptible in code. %
It is possible to compose a series of perturbations $\delta_1, \delta_2, ..., \delta_k$ and apply them to a given code snippet: $\delta_1 \circ \delta_2 ... \circ \delta_k \left(c\right)$.
In this paper, to satisfy the similarity requirement, 
we focus on selecting and applying only a single perturbation.

\paragraph{Semantic-preserving Transformations} 
The advantages of variable renaming as the form of semantic-preserving transformation are that
\begin{inparaenum}[(i)]
\item each variable appears in several places in the code, so a single renaming can induce multiple perturbations;
\item the adversarial code can still be compiled and therefore stays in the code domain;
and \item some variables do not affect the readability of the code and hence, renaming them creates unnoticed unobserved adversarial examples.
\end{inparaenum}

We focus on two distinct types of adversarial examples:
\begin{itemize}
    \item \emph{Targeted attack} -- forces the model to output a specific prediction, which is not the correct prediction.
    \item \emph{Non-targeted attack} -- forces the model to make any incorrect prediction. 
\end{itemize}

From a high-level perspective, the main idea in both kinds of attacks lies in the difference between the standard approach of training a neural network using back-propagation and generating adversarial examples. While training a neural network, we derive the loss with respect to the learned parameters and update each learned parameter. In contrast, when generating adversarial examples, we derive the loss with respect to the \emph{inputs} while \emph{holding the learned parameters constant}, and updating the inputs.

\input{adversarial_descent}
\subsection{Targeted Attack}
\label{subsec:targeted}
In this kind of attack, our goal is to make the model predict an incorrect desired label $y_{bad}$ by renaming a given variable $v$ to $v'$. Let $\theta$ be the learned parameters of a model, $c$ be the input code snippet to the model, $y$ be the target label, and $J(\theta, c, y)$ be the loss function used to train the neural model. As explained in \Cref{subsec:background_training}, when training using gradient descent, the following rule is used to update the model parameters and minimize $J$ :
\begin{equation}
	\theta_{t+1} = \theta_{t} - \eta \cdot \nabla_{\theta} J(\theta_{t}, c, y)
\label{eq:targeted-step}
\end{equation}

We can apply a gradient descent step with $y_{bad}$ as the desired label in the loss function, and derive with respect to any given variable $v$:  
\begin{equation}
    v' = \bar{v} - \eta \cdot \nabla_{\bar{v}} J(\theta, c, y_{bad})
    \label{eq:targeted-step}
\end{equation}

where $\bar{v}$ is the one-hot vector representing $v$.
The above action can be viewed intuitively as taking a step toward the steepest descent, where the direction is determined by the loss function that considers $y_{bad}$ as the correct label. This is illustrated in \Cref{fig:adversarialDescentIllustration} %

The loss $J(\theta, c, y_{bad})$ in
 \Cref{eq:targeted-step} is differentiable with respect to $\bar{v}$ because $J$ is the same loss function that the model was originally trained with, e.g., cross entropy loss. 
 The main difference from the standard gradient descent training (\Cref{eq:gd}) is that \Cref{eq:targeted-step} takes the gradient with respect to the one-hot representation of the input variable ($\nabla_{\bar{v}}$), rather 
 than taking the gradient with respect to the learnable weights ($\nabla_{\theta}$) in \Cref{eq:gd}.

In fact, the result of the above action does not produce a desired new variable name $v'$. Instead it produces a distribution over all possible variable names. To concretize the name of $v'$, we choose the argmax over the resulting distribution, as illustrated in \Cref{onehot} and detailed in \Cref{subsec:deriving}.

\input{ascend_illustration}
\input{onehot_tikz}
\subsection{Non-targeted Attack}
\label{subsec:nontargeted}

In a non-targeted attack, our goal is to update $v$ to $v'$ in a way that will \emph{increase} the loss in \emph{any} direction, instead of decreasing it, as in the training process. Thus, we compute the gradient with respect to $v$ and use Gradient \emph{Ascent}: 
\begin{equation}
    v' = \bar{v} + \eta \cdot \nabla_{\bar{v}} J(\theta, x, y)
    \label{equation:varAsect}
\end{equation}
This rule can be illustrated as taking a step toward the steepest \emph{ascent} in the loss function (\Cref{fig:adversarialAscentIllustration}).

\paragraph{Targeted vs. Non-targeted Attacks} Notice the difference between \Cref{eq:targeted-step} and \Cref{equation:varAsect}.
In \Cref{eq:targeted-step}, the goal of the targeted attack is to find the direction of the \emph{lowest} loss with respect to the \emph{adversarial label}. So, we take a step toward the \emph{negative} gradient.

In \Cref{equation:varAsect}, our goal for the non-targeted attack is to find the direction of the \emph{higher} loss with respect to the \emph{original} label. This is exactly what the gradient gives us.

These equations differ on two cardinal axes: the target label (original or adversarial), and direction of progress (towards or away from), which are the main axes of the equations.

\subsection{Deriving by Integer Indices}
\label{subsec:deriving}
The operation of looking-up a row vector in an embedding matrix using its index is simple. Some common guides and tutorials describe this as taking the dot product of the embedding matrix and a \emph{one-hot} vector representing the index. 
In contrast with these guides, when implementing neural networks, there is usually no real need to use one-hot vectors \emph{at all}. All word embeddings can be stored in a matrix such that each row in the matrix corresponds to a word vector. Looking up a specific vector is then performed by simply looking up a row in the matrix using its \emph{index}. 

Nevertheless, in the context of adversarial examples, deriving the loss with respect to a single variable name is equivalent to \emph{deriving with respect to an index}, which is zero almost everywhere. 
Thus, instead of using indices, we have to represent variable names using one-hot vectors, because these \emph{can} be derived. Looking up a vector in a matrix can then be performed by taking the dot product of the embedding matrix with the one-hot vector. 
Deriving the loss by a one-hot vector instead of an index is thus equivalent to deriving by the (differentiable) \emph{distribution over indices}, rather than deriving by the index itself. 
The result of each \emph{adversarial step} is thus a distribution over all variable names, in which we select the argmax (\Cref{onehot}).

\subsection{Search}
\label{subsec:search}
Sometimes, adversarial examples can be found by applying the \emph{adversarial step} of \Cref{eq:targeted-step} once. At other times, multiple steps are needed, i.e., replacing $v$ with $v'$ and computing the gradient again with respect to $v'$. We limit the number of times we apply the \emph{adversarial step} by a $depth$ hyperparameter. 
Additionally, instead of taking the $argmax$ from the distribution over candidate names, we can try all ``top-k'' candidates.
These define a Breadth-First Search (BFS), where the $width$ parameter is defined by the ``top-k''. Checking whether a choice of variable name results in the desired output is low cost; it just requires computing the output of the model given the modified input. Thus, it is conveniently feasible to perform multiple adversarial steps and check multiple top-k candidates.

The $depth$ and $width$ hyperparameters fine tune the trade-off between the effectiveness of the attack and its computation cost. Small values will define a search that is computationally inexpensive but may lead to early termination without any result. Higher values can increase the probability of finding an adversarial example at the cost of a longer search.

\subsection{Convergence}
\label{subsec:convergence}
In general, gradient descent is guaranteed to converge to a global minimum only when optimizing \emph{convex} functions \cite{nesterov2013introductory}. The loss functions of most neural networks are not convex, and thus, our gradient descent process of multiple adversarial steps is \emph{not} guaranteed to succeed. Furthermore, even if we could hypothetically find a global minimum to the loss function $J(\theta, c, y_{bad})$, it is possible that for a given adversarial target $y_{bad}$ there is no perturbation that can force the model to predict $y_{bad}$. In fact, it is rather intriguing that the success rate for our targeted attack is so high (as show in \Cref{sec:eval}) when we restrict the perturbation to a single variable.

As discussed in \Cref{subsec:search}, we limit the number of steps by a $depth$ hyperparameter. Empirically, in most cases this process ends successfully, by finding a new variable name $v'$ that forces the adversarial label $y_{bad}$. If no adversarial $v'$ was found after $depth$ steps, we consider this attack to have failed.

While recent work provides proofs and guarantees for continuous domains \cite{singh2019abstract,balunovic2019certifying}, we are not aware of any work that provides guarantees for discrete domains.
Even in continuous domains, guarantees have only been provided for small networks with just 88K learnable parameters \cite{singh2019abstract} and 45K parameters \cite{balunovic2019certifying}; these networks often restrict the use of non-linearities. In contrast, the models we experiment with are state-of-the-art, realistic, models with several orders of magnitude more parameters.

%% file: adversarial_descent.tex
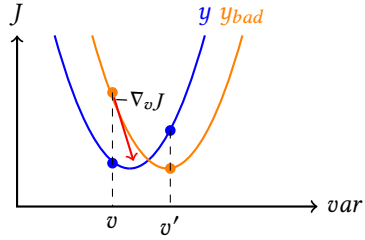
\begin{figure}
\begin{tikzpicture}
	\draw [<->,thick] (0,2.25) node (yaxis) [above] {$J$}
        |- (4,0) node (xaxis) [right] {$var$};
	  \draw[thick,blue] (1.5,0.5) parabola (2.5,2.25) node[above] {$y$};
	  \draw[thick,blue] (1.5,0.5) parabola (0.5,2.25) ;
	  
	  \draw[thick,orange] (2,0.5) parabola (3,2.25) node[above] {$y_{bad}$};
	  \draw[thick,orange] (2,0.5) parabola (1,2.25) ;

	  \coordinate (v) at (1.27,0.57) ;
	  \coordinate (vbad) at (1.27,1.5) ;
	  \coordinate (vtag) at (2.04,1) ;
	  \coordinate (vtagbad) at (2.04,0.5) ;
	  \coordinate (a) at (1.55,0.6) ; 
	  \draw[->,thick,red] (vbad) -- (a) node[black,midway,sloped,above,rotate=70,yshift=1mm,xshift=2.1mm] {{\footnotesize $ -\nabla_{v} J$}};; 
	  \fill[blue] (v) circle (2pt) ;
	  \fill[orange] (vbad) circle (2pt) ;
	  \fill[orange] (vtagbad) circle (2pt) ;
	  \fill[blue] (vtag) circle (2pt) ;
	  \draw[dashed] (vbad) -- (v) -- (xaxis -| v) node[below] {$v$};
	  \draw[dashed] (vtag) -- (xaxis -| vtag) node[below] {$v'$};
\end{tikzpicture}
\caption{Gradient is computed for $y_{bad}$ loss function w.r.t. $v$. By moving toward the opposite direction of the gradient (and replacing $v$ with $v'$), we decrease the loss of $y_{bad}$.}\label{fig:adversarialDescentIllustration}
\end{figure}

%% file: ascend_illustration.tex
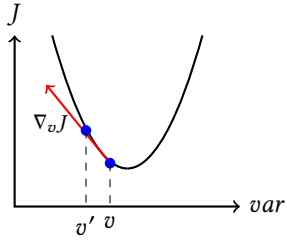
\begin{figure}
\begin{tikzpicture}
	\draw [<->,thick] (0,2.25) node (yaxis) [above] {$J$}
        |- (3,0) node (xaxis) [right] {$var$};
	  \draw[thick] (1.5,0.5) parabola (2.5,2.25) ;
	  \draw[thick] (1.5,0.5) parabola (0.5,2.25) ;
	  \coordinate (v) at (1.27,0.57) ;
	  \coordinate (vtag) at (0.95,1) ;
	  \coordinate (a) at (0.42,1.6) ; 
	  \draw[->,thick,red] (v) -- (a) node[black,midway,sloped,left,rotate=50] {{\footnotesize $ \nabla_{v} J$}};; 
	  \fill[blue] (v) circle (2pt) ;
	  \fill[blue] (vtag) circle (2pt) ;
	  \draw[dashed] (v) -- (xaxis -| v) node[below] {$v$};
	  \draw[dashed] (vtag) -- (xaxis -| vtag) node[below] {$v'$};
\end{tikzpicture}
\caption{Gradient ascent illustration: Gradient is computed with respect to $v$. By moving towards the gradient's direction (and replacing $v$ with $v'$), we increase the loss.}	  \label{fig:adversarialAscentIllustration}
\end{figure}

%% file: onehot_tikz.tex
\begin{figure}
\footnotesize
\input{onehot_tabular.tex}

\caption{Perturbing a variable name: the original variable name is represented as a one-hot vector over the variable-name vocabulary. After perturbation, the vector is no longer one-hot. We apply argmax to find the most likely adversarial name, resulting with another one-hot vector over the variable-name vocabulary. This figure is identical to \Cref{fig:onehot-intro} and repeated here for clarity.}
\label{onehot}
\end{figure}

%% file: defense.tex
\section{Defense Against Adversarial Examples}
\label{sec:defense}

To defend against adversarial attacks, we consider two broad classes of defense techniques. 
The first class contains techniques 
that serve as a ``gatekeeper'' for incoming input and can be plugged in on top of existing trained models, without having to re-train them.
The second class contains techniques that require the model to be re-trained, possibly using a modified loss function or a modified version of the original training set.

\subsection{Defense Without Re-training} 
\label{subsec:without-retraining}
Techniques that do \emph{not} require re-training are appealing because they allow us to separate the optimization of the model from the optimization of the defense, and easily tune the balance between them. Moreover, training neural models is generally computationally expensive; thus, these approaches enable us to perform the expensive training step only once.

Approaches that do not require retraining can be generalized as placing a defensive-model $g$ before the model of code $f_{\theta}$, which is independent of $g$. The goal of $g$ is to fix the given input, if necessary, in a way that will allow $f_{\theta}$ to predict correctly. Mathematically, the new model can be defined as being composed of two models: $f_{\theta} \circ g$.
We assume that the adversary has access to the model $f_{\theta}$ being attacked, but not to the defense model $g$.

We evaluated the following approaches that do not require re-training:

\para{\emph{No Vars}} - a conservative defensive approach that replaces \emph{all} variables to an \scode{UNK} (``unknown'') symbol, only at test time. 
This approach is 100\% robust by construction, but does not leverage variable names for prediction.

\para{\emph{Outlier Detection}} --
 tries to identify an outlier variable name and filter it out by replacing only this variable with \scode{UNK}.
The main idea is that the adversarial variable name is likely to have a low contextual relation to the other, existing, identifiers and literals in code.
We detect outliers by finding an outlier variable in terms of $L_{2}$ distance among the vectors of the existing variable names. Given a code snippet $c \in \mathcal{C}$, 
we define \textbf{$Sym(c)$} as the set of all identifiers and literals existing in $c$. We select the variable $z^{*}$, which is the most distant from the average of the other symbols:
\begin{equation}	
z^{*} = \operatorname*{argmax}_{z \in Var(c)} 
\norm{
\frac{\sum_{v\in Sym\left(c\right),v\neq z} vec\left(v\right)}{\left|Sym\left(c\right)\right|}
 - vec\left(z\right)} 
\end{equation}

We then define a threshold $\sigma$ that determines whether $z^{*}$ is an outlier. If the $L_2$ distance between the vector of $z$ and the average of the rest of the symbols is greater than $\sigma$, then $z$ is replaced with an \scode{UNK} symbol; otherwise, the code snippet is left untouched. 
Practically, the threshold $\sigma$ is tuned on a validation set. It determines the trade-off between the effectiveness of the defense and the accuracy while not under attack, as we evaluate and discuss in \Cref{subsubsec:tradeoff}.

\subsection{Defense With Re-training} 
\label{subsec:retraining}
Ideally, re-training the original model allows us to train the model to be robust while, at the same time, training the model to perform as well as the non-defensive model. Re-training allows the model to be less vulnerable from the beginning, rather than patching a vulnerable model using a separate defense.

We evaluated the following techniques that do require re-training:

\para{\emph{{Train Without Vars}}} -- replaces all variables with an \scode{UNK} symbol \emph{both at training and test time}. This approach is also 100\% robust by construction, as \emph{No Vars} does not require re-training. It is expected to perform better than \emph{No Vars} in terms of F1 because it is trained not to rely on variable names, and to use other signals instead. The downside is that it requires training a model from scratch, while \emph{No Vars} can be applied to an already-trained model.

\para{\emph{{Adversarial Training}}} -- follows \citet{goodfellow2014explaining} and  trains a model on the original training set, while learning to perform the original predictions \emph{and} training on adversarial examples \emph{at the same time}. Instead of minimizing only the expected loss of the original distribution, every example  from the training set $\left(x,y\right) \in T$ contributes both to the original loss $J(\theta, x, y)$ \emph{and} to an adversarial loss $J_{adv}(\theta, x', y)$. Here, $x'$ is a perturbed version of $x$, which was created using a single BFS step of our non-targeted attack (\Cref{subsec:nontargeted}). During training, we minimized $J+J_{adv}$ simultaneously. Note, this method doesn’t change the model complexity but it does increase the training time, making it about three times slower. 

\para{\emph{{Adversarial Fine-Tuning}}} -- follows \citet{hosseini2017limitation} and trains a model for several epochs with the original examples from the training set using the original loss $J(\theta, x, y)$. Once the model is trained for optimal performance, it is fine-tuned on adversarial examples. During fine-tuning, the model is trained for a single iteration over the training set, and only on adversarial versions of each training example, using the adversarial loss $J_{adv}(\theta, x', y)$. The expectation in this approach is to establish the model's  high performance first, and then ensure the model's robustness to adversarial examples because of the recent fine-tuning.

\para{\emph{No Defense, $\mathbf{|vocab|}$=$\mathbf{\{10k,50k,100k\}}$}} -- trains the original model with a smaller vocabulary. Limiting the vocabulary size has the potential to improve robustness by ignoring rare variable names. Names that are observed in the training or test data but are outside the limited vocabulary are replaced with a special \scode{UNK} symbol. This way, the model is expected to consider only frequent names. Because of their frequency, these names will be observed enough times during training  such that their vector representation is more stable.

%% file: eval.tex
\section{Evaluation}
\label{sec:eval}
Our set of experiments comprises two parts: (a) evaluating the ability of \method{} to change the prediction of the downstream classifier for targeted and non-targeted attacks; and (b) evaluating a variety of defense techniques and their ability to mitigate the attacks.

We note that our goal is \emph{not} to perform a robustness evaluation of the attacked models themselves.
For more on evaluating the robustness for models of code, we refer the reader to \citet{ramakrishnan2020semantic} and \citet{bielik2020adversarial}.
Instead, the goal of this section is to evaluate the effectiveness of our proposed targeted and non-targeted attacks, and to evaluate the robustness that different defense techniques provide to a given model.

Our code, data, and trained models are available at \url{https://github.com/tech-srl/adversarial-examples} .

\subsection{Setup}

\subsubsection{Downstream Models}
We evaluated our \method{} attack using three popular architectures as downstream models.
We obtained the trained models from their original authors, who trained the models themselves.

\para{\sname{code2vec}} was introduced by \citet{alon2019code2vec} as a model that predicts a label for a code snippet. The main idea is to decompose the code snippet into AST paths, and represent the entire snippet as a set of its paths. Using large scale training, the model was demonstrated to predict a method name conditioned on the method body. The goal of the attack is to thus change the predicted method name by perturbing the method body. This model takes Java methods as its inputs, and represents variable names using a vocabulary of learned embeddings.

\para{Gated Graph Neural Networks (\sname{GGNN}s)} were introduced by \citet{li2015gated} as an extension to Graph Neural Networks (GNNs) \cite{scarselli2008graph}. Their aim is to represent the problem as a graph, and to aggregate the incoming messages to each vertex with the current state of the vertex using a GRU \cite{cho2014learning} cell. \sname{GGNN}s were later adapted to source code tasks by \citet{allamanis2018learning}, who applied them to the \sname{VarMisuse} task of predicting the correct variable in a given blank slot among all variables in a certain scope. 

\para{\sname{GNN-FiLM}} is a GNN architecture that was recently introduced by \citet{brockschmidt2019gnn}. It differs from prior GNN models in its message passing functions, which compute the ``neural message'' based on both the source and target of each graph edge, rather than just the source as done in previous architectures. \sname{GNN-FiLM} was shown to perform well for both the \sname{VarMisuse} task as well as other graph-based tasks such as protein-protein interaction and quantum chemistry molecule property prediction. The goal of the attack here and in \sname{GGNN}s is to make the model predict the incorrect output variable name; this is done by changing an unrelated input variable name, which is neither the correct variable nor the adversarial variable.

\paragraph{Vocabulary}
In  \sname{code2vec}, we use the original trained model's own vocabulary to search for a new adversarial variable name. 
We derive the adversarial loss with respect to the distribution over all variable names. Thus, the chosen adversarial variable name can be any name from the vocabulary that was used to train the original \sname{code2vec} model.
The \sname{GNN-FiLM} and the \sname{GGNN} models take C\# methods as their inputs and represent variable names using a \emph{character-level} convolution. Thus, in our \method{} attack we derive the loss with respect to the distribution over all \emph{characters} in a given name; the chosen adversarial input name can be any combination of characters.

\subsubsection{Adversarial Strategies} 
\label{subsubsec:strategies}
While there are a variety of possible adversarial perturbations, we focus on two main adversarial strategies:
\begin{itemize}
    \item \textbf{Variable Renaming (\varname)}: choose a single variable, and iteratively change its name until the model's prediction is changed, using a BFS (as explained in \Cref{sec:approach}). 
Eventually, the ``distance'' between the original code and the adversarial code is \emph{a single variable name at most}.
    \item \textbf{Dead-Code Insertion (\deadcode)}: insert a new unused variable declaration and derive the model with respect to its name. The advantage of this strategy is that the existing code remains unchanged, which might make this attack more difficult to notice. Seemingly, this kind of attack can be mitigated by removing unused variables before feeding the code into the trained model. Nonetheless, in the general case, detecting an unreachable code is undecidable. 
    In all cases, we arbitrarily placed the dead code at the end of the input method, and used our attack to find a new name for the new (unused) declared variable. In our preliminary experiments we observed that placing the dead code \emph{anywhere} else works similarly. In this attack strategy, \emph{a single variable declaration} is inserted. Thus, the ``distance'' between the original code and the adversarial code is \emph{a single variable declaration statement}.

\end{itemize}

Other semantic-preserving transformations such as statement swapping and operator swapping, are \emph{not differentiable} and thus do not enable \emph{targeted attacks}. 

When renaming variables or introducing new variables, we verified that the new variable name does not collide with an existing variable that has the same name.
In the \sname{code2vec} experiments, we used the \emph{adversarial step} %
to run a BFS with $width = 2$ and $depth = 2$. 
In \sname{GGNN} we used $width = 1$ and $depth = 3$, and \sname{GNN-FiLM} required $depth = 10$ to achieve attack success that is close to that of the other models. We discuss these differences in \Cref{subsubsec:attackResults}.
Increasing the $width$ and $depth$ can definitely improve the adversary's success but at the cost of a longer search, although the entire BFS takes a few seconds at most.

\subsubsection{Dataset} 
\label{subsub:dataset}
\para{\sname{code2vec} - Java} We evaluated our proposed attack and defense on \sname{code2vec} using the Java-large dataset \cite{alon2019code2seq}. This dataset contains more than 16M Java methods and their labels, taken from 9500 top-starred Java projects in GitHub that have been created since January 2007. It contains 9000 projects for training, 200 distinct projects for validation, and 300 distinct projects for test. We filtered out methods with no local variables or arguments, since they cannot be perturbed by variable renaming.
To evaluate the effectiveness of our \method{} attack, we focus on the examples that the model predicted correctly out of the test set of Java-large. 

That is, from the original test set, we used a subset that the original \sname{code2vec} predicted accurately.
On this filtered test set, the accuracy of the original \sname{code2vec} model is 100\% by construction.

\para{\sname{GGNN} and \sname{GNN-FiLM} - C\#} We evaluated \method{} with the \sname{GGNN} and \sname{GNN-FiLM} C\# models on the dataset of \citet{allamanis2018learning}. This dataset consists of about 220,000 graphs (examples) from 29 top-starred C\# projects on GitHub. 
For all examples, there is at least one type-correct replacement variable other than the correct variable, and a maximum of up to 5 candidates. \method{} always attacks by modifying an unrelated variable, not the correct variable or the adversarial target. 
In targeted attacks, we randomly pick one out of the five candidates as the target for attack. In non-targeted attacks, our attacker tries to make the model predict any incorrect candidate.
Similar to the Java dataset, we focus on the examples that each original model predicts correctly out of the test set.

\subsection{Attack}
We focus on two main attack tasks: targeted and non-targeted attacks.
For targeted attacks, we used \Cref{eq:targeted-step} as the \emph{adversarial step}.
For non-targeted attacks, we used \Cref{equation:varAsect} as the \emph{adversarial step}. 
For the desired adversarial labels, we randomly sampled labels that occurred at least 10K times in the training set. 

\subsubsection{Metrics}
We measured the robustness of each setting for each of the different attack approaches. The lower model robustness, the higher effectiveness of the attack. 

In \emph{targeted attacks}, the goal of the adversary is to change the prediction of the model to a label of the attacker's desire. We thus define robustness as the percentage of examples in which the correctly predicted label \emph{was not changed to the adversary's desired label}. If the predicted label was changed to a label that is not the adversarial label, we consider the model as robust to the targeted attack.

In \emph{non-targeted attacks}, the goal of the adversary is to change the prediction of the model to any label other than the correct label. We thus define robustness as the percentage of examples in which the correctly predicted label was not changed to any label other than the correct label. 

\subsubsection{Baselines}

Since our task is new, we are not aware of existing baselines. We thus compare \method{} to different approaches in targeted and non-targeted attacks for \sname{code2vec} and GNN models.

\para{\emph{TFIDF}}
is a statistical baseline for attacking \sname{code2vec}: for every pair of a label and a variable name it computes the number of times the variable appears in the training set under this label, divided by the total number of occurrences of the variable in the training set. Then, given a desired adversarial label $y_{bad}$, TFIDF outputs the variable name $v$ that has the highest score with $y_{bad}$:
	$\text{TFIDF}\left(y_{bad}\right)=argmax_{v}\frac{\#\left(y_{bad},v\right)}{\#\left(v\right)}$.

\para{\emph{CopyTarget}}
attacks \sname{code2vec} with \emph{targeted attacks}. \emph{CopyTarget} replaces a variable with the desired adversarial label. For example, if the adversarial label is \scode{badPrediction}, \emph{CopyTarget} renames a variable to \scode{badPrediction}. 

\para{\emph{CharBruteForce}}
attacks \sname{GGNN} and \sname{GNN-FiLM}, which address the \sname{VarMisuse} task. \emph{CharBruteForce} changes the name of the attacked variable by randomly changing every character iteratively up to a limited number of iterations.

\para{\emph{RandomVar}} is used in \emph{non-targeted attacks} on \sname{code2vec}: \emph{RandomVar}  replaces the given variable with a randomly selected variable name from the training set vocabulary. 

In all experiments, the baselines were given the same number of trials as our attack. In \emph{RandomVar}, we randomly sampled the same number of times. In \emph{TFIDF}, we used the top-k TFIDF candidates as additional trials. In \emph{CopyTarget}, we took the target as the new variable name and its k-nearest neighbors in the embedding space.

\input{targets_table}

\subsubsection{Attack - Results} 
\label{subsubsec:attackResults}
\para{\sname{code2vec}} \Cref{table:adversarialOnGoodVsTrivialVsDefense} summarizes the results of \method{} on \sname{code2vec}. The main result is that \method{} outperforms the baselines by a large margin in both targeted and non-targeted attacks.

In non-targeted attacks, \method{} attacks are much more effective than the baseline: for \varname{}, \sname{code2vec} is 6\% robust to \method{}, 34.10\% robust to \emph{RandomVar} and 53.53\% robust to \emph{TFIDF}. For \deadcode, \sname{code2vec} is 21.83\% robust to  \method{}, 54.90\% robust to \emph{RandomVar}, and 84.00\% robust to \emph{TFIDF}. Thus, \method{}'s attack is more effective than that of the baselines.

In targeted attacks, \method{} performs better than \emph{CopyTarget} for each of the randomly sampled adversarial labels.
For example, \sname{code2vec} is only $10.39\%$ robust to \method{} attacks that change the prediction to the label \scode{mergeFrom}, and $72.79\%$ robust to the \emph{CopyTarget} attack of the same target adversarial label.

However, in some of the randomly sampled targets \emph{TFIDF} was more effective than \method{}. In \varname{}, \method{} was the most effective in 6 out of the 10 randomly sampled adversarial targets, while \emph{TFIDF} was the most effective in the remaining 4. In \deadcode{} attacks, \method{} was the most effective in 8 out of 10 randomly sampled adversarial targets, while \emph{TFIDF} was the most effective in the remaining 2. 
Although \emph{TFIDF} performed better than \method{} for a few of the targets, the differences were usually small. 
Overall, \method{}'s targeted attack is more effective than that of the baselines.

Non-targeted attacks generally yield lower model robustness than targeted attacks. This is expected, since non-targeted attacks try to change the label to \emph{any} label other than the correct one, while targeted attacks count as successful only if they change the prediction to the desired label.

In general, the \varname{} attack is more effective than \deadcode. We hypothesize that this is because the inserted unused variable impacts only a small part of the code. Hence, it may have a smaller numerical effect on the computation of the model. In contrast, renaming an existing variable changes multiple occurrences in the code and thus has a wider effect.

\input{graphmodel_targets_table}

\para{\sname{GGNN} and \sname{GNN-FiLM}} \Cref{table:graphModelsadversarialVsTrivial} summarizes the results of the \method{} attack and the baseline on the \sname{GGNN} and \sname{GNN-FiLM} models.
The main result is that \method{} is much more effective than the \emph{CharBruteForce} baseline. 
The \sname{GGNN} model is $69.00\%$ robust to the \method{} targeted attack, and  $98.84\%$ robust to the baseline attack. The \sname{GNN-FiLM}  is $87.62\%$ robust to the \method{} attack and $96.19\%$ robust to the \emph{CharBruteForce} attack. 

In addition, we see that \method{} is more effective on \sname{code2vec} than on both GNN architectures. \sname{code2vec} is $6\%$ robust to non-targeted \varname{} attacks and \sname{GGNNs} is $57.99\%$ robust to non-targeted attacks. There are several reasons for this difference between the attacked models: 
\begin{enumerate}[]
\item \sname{code2vec} is simpler and more ``neurally shallow'', while \sname{GGNN} uses 6 layers of message propagation steps, and \sname{GNN-FiLM} uses 10 layers.
\item The models' tasks are very different: \sname{code2vec} classifies a given code snippet to 1 out of about 200,000 possible target method names, while both GNN architectures address the \sname{VarMisuse} task and need to choose 1 out of only 2 to 5 possible variables.
\item \sname{code2vec} has orders of magnitude more trainable parameters (about 350M) than \sname{GGNN} (about 1.6M) and \sname{GGNN} (about 11.5M), making \sname{code2vec} more sparse, its loss hyperspace more complex, and thus more vulnerable.
\end{enumerate}

Finally, we see that the \sname{GNN-FiLM} model is more robust than \sname{GGNN}. Even though we used a more aggressive BFS with 10 gradient steps for \sname{GNN-FiLM} and only 3 gradient steps for \sname{GGNN}, the \sname{GNN-FiLM} robustness to \method{} targeted attack is $87.62\%$, while the \sname{GGNN} robustness is $69.00\%$. We hypothesize that this is primarily because the message passing function in \sname{GNN-FiLM} computes the sent message between nodes based on both the source \emph{and target} of each graph edge, rather than just the source as in the \sname{GGNN}. This makes it more robust to an attack on a single one of these nodes. 
The higher results of \sname{GNN-FiLM} over \sname{GGNN} in other graph-based benchmarks \cite{brockschmidt2019gnn} hint that \sname{GNN-FiLM} may be using the graph topology, i.e., the program structure, to a larger degree; the \sname{GGNN} focuses mainly on the names and is thus more vulnerable to name-based attacks.
We leave further investigation of the differences between different GNN architectures for future work.

\input{summarize_table}

\subsection{Defense}

We experimented with all defense techniques
as described in \Cref{sec:defense}, in a \sname{code2vec} model. 
\emph{No Defense} is the vanilla, unmodified, trained model that was trained by its original authors. \emph{No Defense, $|vocab|$=$\{10k,50k,100k\}$} are vanilla code2vec models that we trained using the authors' code and the default settings, without any special defense or modification, except for limiting the vocabulary sizes (where vocabulary size is mentioned); this allowed us to examine whether the limited vocabulary size can serve as some sort of defense.

\subsubsection{Metrics} %
We measured the success rate of the different defense approaches in preventing the adversarial attack and increasing robustness.
When evaluating alternative defenses, it is also important to measure the performance of the original model while using the defense, \emph{but not under attack}. An overly defensive approach can lead to $100\%$ robustness at the cost of reduced prediction performance.

To tune the threshold $\sigma$ of the \emph{Outlier Detection} defense, we balanced the following factors: 
\begin{inparaenum}[(1)]
\item the robustness of the model using this defense;
and 
\item the F1 score of the model using this defense, while not under attack. 
\end{inparaenum}
We tuned the threshold on the validation set, and chose $\sigma = 2.7$ since it led to $75\%$ robustness against non-targeted attack at the cost of $2\%$ degradation in F1 score while not under attack. However, this threshold can be tuned according to the desired needs in the trade-off between performance and defense (see \Cref{subsubsec:tradeoff}).

\input{defense-per-target-table}
\input{defenses-scatter.tex}
\subsubsection{Defense - Results} 
The effectiveness of the different defense techniques is presented in \Cref{table:OriginalAndAdversarialResultsOnGoodCompareGuard}. 
The main results are as follows: \emph{Outlier Detection} provides the best performance and highest robustness among the techniques that do not require re-training; it achieves an F1 of 97.02 and above 75.35\% robustness for targeted and non-targeted attacks. Among the techniques that do require re-training, \emph{Adversarial Training} achieves the highest performance and robustness; \emph{Adversarial Training} achieves an F1 score of 94.98 and above 68.80\% robustness for targeted and non-targeted attacks.

The penalty of using \emph{Outlier Detection} is only $2.18\%$ degradation in F1 score compared to \emph{No Defense}. At the other extreme, \emph{Train Without Vars} is $100\%$ robust to \varname{} and \deadcode{} attacks, but its F1 is degraded by $9.6\%$ compared to \emph{No Defense}. That is, \emph{Outlier Detection} is a sweet spot in the trade-off between performance and robustness: it is $98\%$ as accurate as the original model (\emph{No Defense}) and $74-96\%$ as robust as the model that ignores variables completely.

\emph{Adversarial Training} provides a sweet-spot among the techniques that do require re-training: its penalty is $5.02\%$ degradation in F1 score compared to \emph{No Defense}, and it achieves over $99\%$ robustness to \deadcode{} and $94.48\%$ robustness to targeted \varname. Adversarial training allows the model \emph{to leverage} variable names to achieve high performance while not under attack, but also to not overly rely on variable names; it is thus robust to adversarial examples.
\emph{Adversarial Fine-Tuning} performs surprisingly worse than \emph{Adversarial Training}, both in F1 score and in its robustness.

\emph{No Vars} is 100\% robust as \emph{Train Without Vars}, but is about 10 F1 points worse than \emph{Train Without Vars}. The only benefit of \emph{No Vars} over \emph{Train Without Vars} is that \emph{No Vars} can be applied to a model without re-training it.

Reducing the vocabulary size to 100K and 50K, results in a negligible decrease in performance while not under attack, compared to not defending at all. However, it also results in roughly the same poor robustness as no defense at all.
Reducing the vocabulary size to 10K hurts performance while not under attack and does not provide much robustness.

These results are visualized in \Cref{fig:defenses-scatter}.

\Cref{table:defense-per-target} shows the performance of the \emph{Outlier Detection} and the \emph{Adversarial Training} defenses across randomly sampled adversarial labels.

\input{tradeoff-fig}
\subsubsection{Robustness - Performance Trade-off}
\label{subsubsec:tradeoff}
One of the advantages of the \emph{Outlier Detection} defense, which we found to be one of the most effective defense techniques, is the ability to tune its trade-off between robustness and performance. \Cref{fig:vulnerabilityF1Tradeoff} shows the trade-off with respect to the similarity threshold $\sigma$. 
Small values of $\sigma$ make the model with \emph{Outlier Detection} defense more robust. It becomes almost as robust as \emph{No Vars}, but perform worse in terms of F1 score. As the value of $\sigma$ increases, the model with \emph{Outlier Detection} defense becomes less robust, but performs better while not under attack.

%% file: targets_table.tex
\begin{table*}[]
\small
\begin{tabular}{ l  c c c  c c c  }
\toprule
 & \multicolumn{3}{c}{\varname{} (robustness \%)} & \multicolumn{3}{c}{\deadcode{} (robustness \%)} \\ 
 \midrule
 &  \underline{\emph{RandomVar}} & \underline{TFIDF} & \underline{\method{} (this work)} & \underline{\emph{RandomVar}} & \underline{TFIDF} & \underline{\method{} (this work)}  \\
Non-targeted  & 34.10 & 53.53 & \textbf{6.00} & 54.90 & 84.00 & \textbf{21.83}  \\ 
\midrule
\midrule
Targeted: & \underline{\emph{CopyTarget}} &  &   & \underline{\emph{CopyTarget}} & &  \\
init & 84.47 & 74.33 & \textbf{48.44} & 96.79 & 96.67 & \textbf{87.62}  \\
mergeFrom & 72.79 & \textbf{9.01} & 10.39 & 99.82 & 29.34 & \textbf{22.65} \\
size & 99.47 & \textbf{77.91} & 78.27 & 99.97 & 98.98 & \textbf{95.96} \\
isEmpty & 88.61 & 98.11 & \textbf{79.04} & 99.98 & 99.72 & \textbf{87.63}  \\
clear & 89.56 & 89.00 & \textbf{82.80} & 99.07 & 98.55 & \textbf{97.89}  \\
remove & 84.94 & 84.83 & \textbf{63.15} & 99.29 & 99.24 & \textbf{80.02}  \\
value & 99.77 & \textbf{71.81} & 76.75 & 100.0 & \textbf{98.16} & 98.33  \\
load & 86.75 & 85.46 & \textbf{55.65} & 99.03 & 97.22 & \textbf{86.65}  \\
add & 92.88 & 86.72 & \textbf{68.60} & 99.93 & 95.29 & \textbf{93.75}  \\
run & 95.11 & \textbf{40.92} & 51.52 & 99.36 & \textbf{62.87} & 77.63  \\ 
\midrule
Count best: & 0 & 4 & \textbf{6} & 0 & 2 & \textbf{8} \\ 
\bottomrule
\end{tabular}
\caption{Robustness of \sname{code2vec} to our adversarial attacks, targeted and non-targeted, using \varname{} and \deadcode{}, compared to the baselines (the lower robustness, the more effective the attack). \method{} is more effective than the baselines in 6 out of 10 \varname{} randomly sampled targets and in 8 out of 10 \deadcode{} targets.}
\label{table:adversarialOnGoodVsTrivialVsDefense}
\end{table*}

%% file: graphmodel_targets_table.tex
\begin{table*}[]
\small
\begin{tabular}{ l  c c c c c  }
\toprule
 & \multicolumn{2}{c}{\sname{\sname{GGNN}} Robustness} & & \multicolumn{2}{c}{\sname{GNN-FiLM} Robustness} \\
 \cmidrule{2-3}  \cmidrule{5-6}
 & \emph{CharBruteForce} & \method{} (this work) &  & \emph{CharBruteForce} & \method{} (this work)  \\ 
 \midrule
Non-targeted & 96.24 & \textbf{57.99}  & & 95.56 & \textbf{83.55}  \\ 
targeted & 97.84 & \textbf{69.00} &  & 96.19 & \textbf{87.62}  \\ 
\bottomrule
\end{tabular}
\caption{Robustness of \sname{GGNN} and \sname{GNN-FiLM} to our adversarial attacks (targeted and non-targeted), compared to \emph{CharBruteForce} (the lower the robustness, the more efficient the attack). Our attack is more effective than brute-force given an equal number of trials. Every successful attack results in a type-safe \sname{VarMisuse} bug.}
\label{table:graphModelsadversarialVsTrivial}
\end{table*}

%% file: summarize_table.tex
\begin{table*}[t]
\footnotesize
\begin{tabular}{lccccccc}
\toprule
           & \multicolumn{3}{c}{Performance} & \multicolumn{2}{c}{\varname{}} & \multicolumn{2}{c}{\deadcode{}}   \\
           & \multicolumn{3}{c}{(not under attack)} & \multicolumn{2}{c}{Robustness (\%) } & \multicolumn{2}{c}{Robustness (\%) }  \\
\cmidrule{2-4} \cmidrule{5-6} \cmidrule{7-8}
           & Prec & Rec & F1 & Non-targeted & Target: ``run'' & Non-targeted & Target: ``run'' \\
           \midrule
No Defense & \textbf{100} & \textbf{100} & \textbf{100} & 6.0 & 51.52 & 22.83 & 77.63 \\
No Vars & 78.78 & 80.83 & 79.98 & \textbf{100} & \textbf{100} & \textbf{100} & \textbf{100} \\
Outlier Detection & \textbf{98.18} & \textbf{97.75} & \textbf{97.92} & 74.35 & \textbf{96.49} & \textbf{96.73} & \textbf{95.76} \\
Train Without Vars & 89.74 & 90.86 & 90.40 & \textbf{100} & \textbf{100} & \textbf{100} & \textbf{100} \\
Adversarial Training & \textbf{98.09} & 92.96 & 94.98 & 68.80 & 94.48 & \textbf{99.33} & \textbf{99.98} \\
Adversarial Fine-Tuning & 85.51 & 84.86 & 85.12 & 13.59 & 68.94 & 31.07 & 56.64 \\
No Defense, $|vocab|$=$10k$ & 90.50 & 94.26 & 92.66 & 6.56 & 45.78 & 31.13 & 94.55 \\
No Defense, $|vocab|$=$50k$ & 94.51 & \textbf{100} & \textbf{98.00} & 3.18 & 40.48 & 20.01 & 94.04 \\
No Defense, $|vocab|$=$100k$ & \textbf{96.84} & \textbf{99.97} & \textbf{98.99} & 2.61 & 32.48 & 19.64 & 68.58 \\
\bottomrule
\end{tabular}
\caption{
Precision, recall, F1, and robustness percentage of different models. 
The higher the robustness, the more effective the defense. 
Scores that are above 95\% are marked in \textbf{bold}.
Precision, recall, and F1 are measured on the subset of the test set that the vanilla model predicts accurately (as explained in \Cref{subsub:dataset}).
\emph{Outlier Detection} and \emph{Adversarial Training} are sweet-spots: they perform almost as good as \emph{No Defense} in terms of precision, recall, and F1, and they are almost as robust as the extreme \emph{No Vars}.}
\label{table:OriginalAndAdversarialResultsOnGoodCompareGuard}
\end{table*}

%% file: defense-per-target-table.tex
\begin{table*}[]
\small
\begin{tabular}{ l  c c c c c c c }
\toprule
 &  \multicolumn{3}{c}{\varname{} (robustness \%)} & & \multicolumn{3}{c}{\deadcode{} (robustness \%)} \\
\cmidrule{2-4} \cmidrule{6-8}
 &  \method$^{\dagger}$ & \makecell{\method{} \\ + \emph{Outlier} \\ \emph{Detection}} & \makecell{\method{} \\ + \emph{Adversarial} \\ \emph{Training}} &  & \method{}$^\dagger$ & \makecell{\method{} + \\ \emph{Outlier} \\ \emph{Detection}} & \makecell{\method{} \\ + \emph{Adversarial} \\ \emph{Training}} \\ 
 \midrule
init &  48.44 & 74.32 & \textbf{78.17} &  & 87.62 & 91.41 & \textbf{99.97} \\
mergeFrom &  10.39 & \textbf{99.98} & 98.91 &  & 22.65 & \textbf{99.99} &  \textbf{100.00} \\
size &  78.27 & \textbf{99.58} & 99.39 &  & 95.96 & 99.94  & \textbf{99.99} \\
isEmpty & 79.04 & \textbf{99.22} & 98.91 &  & 87.63 & 97.03  & \textbf{99.99}  \\
clear &  82.8 & \textbf{98.77} & 85.09 &  & 97.89 & 99.56  & \textbf{99.99}\\
remove &  63.15 & \textbf{94.50} & 89.29 &  & 80.02 & 99.33  & \textbf{99.99} \\
value &  76.75 & 90.87 & \textbf{99.47} &  & 98.33 & 99.72  & \textbf{100.00}\\
load &  55.65 & 60.27 & \textbf{88.29} &  & 86.65 & 85.28  & 100.00 \\
add &  68.6 & 88.97 & \textbf{95.90} &  & 93.75 & 97.69  & \textbf{100.00}\\
run &  51.52 & \textbf{96.49} & 94.48 &  & 77.63 & 95.76  & \textbf{99.98} \\ 
\bottomrule
\end{tabular}
\caption{Robustness of \sname{code2vec} to adversarial attacks with the \emph{Outlier Detection} and the \emph{Adversarial Training} defenses, across different adversarial targets (the higher the robustness --- the more effective the defense). 
\method$^{\dagger}$ results are the same as in \Cref{table:adversarialOnGoodVsTrivialVsDefense}.}
\label{table:defense-per-target}
\end{table*}

%% file: defenses-scatter.tex
\begin{figure}[h!]
\begin{tikzpicture}[scale=1.0]
	\begin{axis}[
		xlabel={Average Robustness (\%)},
		ylabel={Performance \\ (F1)},
        legend style={at={(0.97,0.2)},anchor=east,font=\tiny},
        xmin=0, xmax=101,
        ymin=75, ymax=101,
        xtick={0,10,...,100},
        ytick={0,5,...,100.0},
		ylabel style={rotate=-90, align=center,at={(-0.05,0.5)}},
		grid = major,
        major grid style={dotted,black},
        width = 0.8\linewidth, height = 6cm
    ]
	\addplot[
    blue,
    thick,
    mark=star, only marks,
    mark options={fill=white},
    visualization depends on=\thisrow{alignment} \as \alignment,
    nodes near coords, %
    point meta=explicit symbolic, %
    every node near coord/.style={anchor=\alignment} %
    ] table [%
     meta index=2 %
     ] {
x       y       label       alignment
39.495  100    {No Defense}      180
100    79.98    {No Vars}          0
90.832    97.92    \textbf{Outlier Detection}          -30
};

	\addplot[
    blue,
    thick,
    mark=*, only marks, mark size=1.5pt,
    mark options={fill=blue},
    visualization depends on=\thisrow{alignment} \as \alignment,
    nodes near coords, %
    point meta=explicit symbolic, %
    every node near coord/.style={anchor=\alignment} %
    ] table [%
     meta index=2 %
     ] {
x       y       label       alignment
44.505    92.66    {$|vocab|$=$10k$}          90
39.428    98.00    {$|vocab|$=$50k$}          90
30.8275    98.99    {$|vocab|$=$100k$}          0
100    90.40    {Train Without Vars}          0
90.648    94.98    {\textbf{Adversarial Training}}          0
42.56    85.12    {Adversarial Fine-Tuning}          90
};

	\end{axis}

\end{tikzpicture}
    \caption{The performance of each defense technique compared to its average robustness. Average robustness is calculated as the average of the four robustness scores in \Cref{table:OriginalAndAdversarialResultsOnGoodCompareGuard}.
    Points marked with a bullet ($\bullet$) require re-training; points marked with a star ($\star$) do not require re-training. \emph{Outlier Detection} provides the highest performance and robustness among the techniques that do \emph{not} require re-training; \emph{Adversarial Training} provides the highest performance and robustness among the techniques that \emph{do} require re-training.}
    \label{fig:defenses-scatter}
\end{figure}
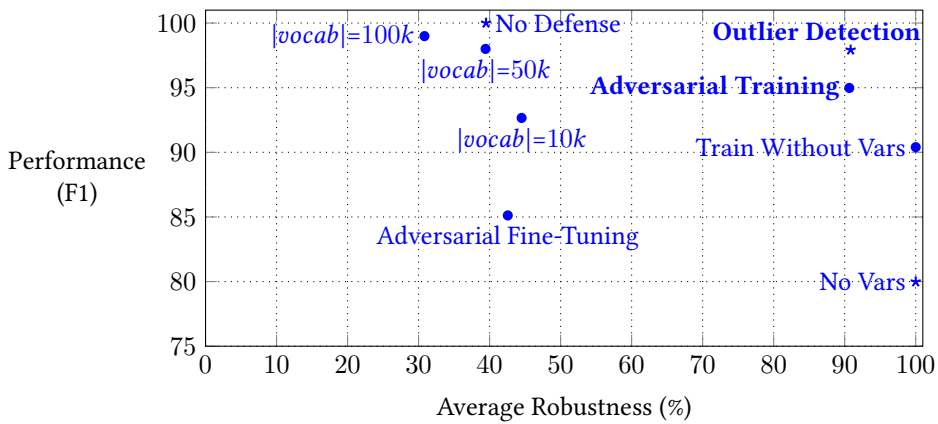 

%% file: tradeoff-fig.tex
\begin{figure*}[]
\begin{tikzpicture}[scale=0.9]
	\begin{axis}[
		xlabel={Threshold ($\sigma$)},
		ylabel={F1 score},
		ylabel near ticks,
        legend style={at={(0.99,0.4)},anchor=east},
        xmin=0, xmax=6,
        ymin=80, ymax=101,
        xtick={0,...,6},
        ytick={80,85,...,100},
        grid = major,
        major grid style={dotted,black},
        width = 0.8\linewidth, height = 5cm
    ]
    \addplot[color=red,line width=1pt] coordinates {
	(0.0,100.0)
	(6.0,100.0)
	}; 
    \addlegendentry{No Defense}

    \addplot[color=blue,line width=1.5pt] coordinates {
	(0.0,86.37921239629318)
	(0.1,86.37921239629318)
	(0.2,86.37921239629318)
	(0.3,86.37921239629318)
	(0.4,86.37921239629318)
	(0.5,86.37921239629318)
	(0.6,86.37921239629318)
	(0.7,86.37921239629318)
	(0.8,86.37921239629318)
	(0.9,86.37921239629318)
	(1.0,86.37921239629318)
	(1.1,86.3794852755858)
	(1.2,86.38643833983716)
	(1.3,86.40610042222698)
	(1.4,86.42434823976441)
	(1.5,86.49314796896948)
	(1.6,87.36583675342924)
	(1.7,87.66717648120398)
	(1.8,88.19025924473961)
	(1.9,88.97088522301506)
	(2.0,89.90367963473051)
	(2.1,90.69312038883729)
	(2.2,91.93895020648861)
	(2.3,93.18779725622771)
	(2.4,94.54036303403109)
	(2.5,95.74076088696437)
	(2.6,97.08225043568316)
	(2.7,97.93596113114697)
	(2.8,98.54633799955454)
	(2.9,98.98790510963099)
	(3.0,99.26326442306313)
	(3.1,99.47500388983303)
	(3.2,99.62702894174707)
	(3.3,99.70797978493142)
	(3.4,99.77112348414494)
	(3.5,99.81310838778403)
	(3.6,99.84409396987967)
	(3.7,99.87603918967227)
	(3.8,99.89860305789134)
	(3.9,99.9156381769827)
	(4.0,99.93749600694196)
	(4.1,99.96568442298158)
	(4.2,99.97411369352159)
	(4.3,99.98565510284735)
	(4.4,99.98763007106561)
	(4.5,99.98773594810605)
	(4.6,99.98773594810605)
	(4.7,99.98960503510125)
	(4.8,99.98960503510125)
	(4.9,99.99210940219065)
	(5.0,99.99314982702342)
	(5.1,99.99688797808285)
	(5.2,99.99740818829582)
	(5.3,100.00010588946624)
	(5.4,100.00010588946624)
	(5.5,100.00010588946624)
	(5.6,100.0)
	(5.7,100.0)
	(5.8,100.0)
	(5.9,100.0)
	(6.0,100.0)
	}; 
    \addlegendentry{\textbf{\emph{Outlier Detection}}}

    \addplot[color=OliveGreen,line width=1pt] coordinates {
	(0.0,81.06671541752713)
	(6.0,81.06671541752713)
	}; 
    \addlegendentry{No Vars}
	
	\addplot[color=orange,line width=1pt] coordinates {
	(0.0,91.86653179074791)
	(6.0,91.86653179074791)
	}; 
    \addlegendentry{Train Without Vars}
    
   	\addplot[color=Plum,line width=1pt] coordinates {
	(0.0,93)
	(6.0,93)

	}; 
    \addlegendentry{Adversarial Training}
    
    \addplot[color=black,line width=0.5pt, style=dashed] coordinates {
	(2.7,80)
	(2.7,110)
	};
    
	\end{axis}

\end{tikzpicture}

\begin{tikzpicture}[scale=0.9]
	\begin{axis}[
		xlabel={Threshold ($\sigma$)},
		ylabel={Robustness (\%)},
		ylabel near ticks,
        legend style={at={(0.99,0.6)},anchor=east},
        xmin=0, xmax=6,
        ymin=0, ymax=105,
        xtick={0,...,6},
        ytick={0,20,...,100},
        grid = major,
        major grid style={dotted,black},
        width = 0.8\linewidth, height = 5cm
    ]
    \addplot[color=red,line width=1pt] coordinates {
	(0.0,4.16)
	(6.0,4.16)
	}; 
    \addlegendentry{No Defense}

    \addplot[color=blue,line width=1.5pt] coordinates {
	(0.0,95.67)
	(0.1,95.66)
	(0.2,95.67)
	(0.3,95.66)
	(0.4,95.66)
	(0.5,95.66)
	(0.6,95.67)
	(0.7,95.67)
	(0.8,95.66)
	(0.9,95.67)
	(1.0,95.66)
	(1.1,95.66)
	(1.2,95.65)
	(1.3,95.62)
	(1.4,95.57)
	(1.5,95.42)
	(1.6,92.97)
	(1.7,92.3)
	(1.8,91.07)
	(1.9,89.53)
	(2.0,87.94)
	(2.1,86.74)
	(2.2,85.18)
	(2.3,83.41)
	(2.4,81.85)
	(2.5,80.27)
	(2.6,77.97)
	(2.7,75.12)
	(2.8,70.89)
	(2.9,66.19)
	(3.0,61.29)
	(3.1,56.330000000000005)
	(3.2,51.910000000000004)
	(3.3,47.97)
	(3.4,44.089999999999996)
	(3.5,40.690000000000005)
	(3.6,37.69)
	(3.7,35.28)
	(3.8,32.989999999999995)
	(3.9,30.75)
	(4.0,28.580000000000013)
	(4.1,26.239999999999995)
	(4.2,23.760000000000005)
	(4.3,21.599999999999994)
	(4.4,19.86)
	(4.5,18.450000000000003)
	(4.6,16.629999999999995)
	(4.7,14.959999999999994)
	(4.8,13.409999999999997)
	(4.9,11.989999999999995)
	(5.0,10.900000000000006)
	(5.1,9.579999999999998)
	(5.2,8.420000000000002)
	(5.3,7.260000000000005)
	(5.4,6.489999999999995)
	(5.5,5.910000000000011)
	(5.6,5.47999999999999)
	(5.7,5.150000000000006)
	(5.8,4.849999999999994)
	(5.9,4.540000000000006)
	(6.0,4.240000000000006)
	}; 
    \addlegendentry{\textbf{\emph{Outlier Detection}}}

    \addplot[color=OliveGreen,line width=1pt] coordinates {
	(0.0,100)
	(6.0,100)
	}; 
    \addlegendentry{No Vars}
	
	\addplot[color=orange,line width=1pt] coordinates {
	(0.0,99.3)
	(6.0,99.3)
	}; 
    \addlegendentry{Train Without Vars}
    
    \addplot[color=Plum,line width=1pt] coordinates {
	(0.0,69.84)
	(6.0,69.84)
	}; 
    \addlegendentry{Adversarial Training}
    
	\addplot[color=black,line width=0.5pt, style=dashed] coordinates {
	(2.7,0)
	(2.7,110)
	};
    
	\end{axis}

\end{tikzpicture}

\caption{The \emph{Outlier Detection} defense: the trade-off between robustness and performance-while-not-under-attack, with respect to the similarity threshold $\sigma$ on the validation set. 
The dashed vertical line at $\sigma=2.7$ denotes the value that we chose for $\sigma$ according to the validation set.
A lower threshold leads to perfect robustness and lower performance; a higher threshold leads to performance that is equal to the original model's, but the model is also as vulnerable as the original model. The robustness score is the robustness against non-targeted \varname{} attacks.}
\label{fig:vulnerabilityF1Tradeoff}
\end{figure*}
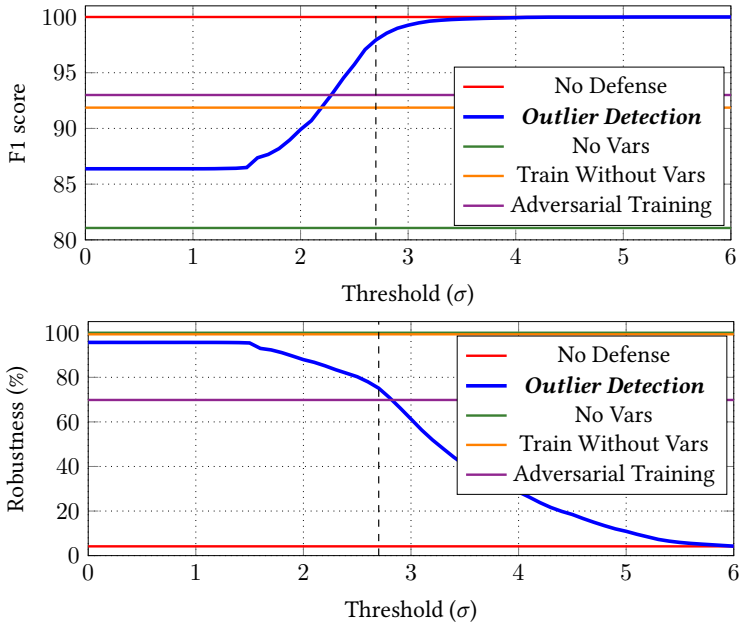 

%% file: eval_examples.tex
\input{example_sort}

\input{example_deadcode}

\input{example_property_short.tex}

\input{example_gnn_destinationType.tex}

\subsection{Additional Examples}
All examples that are shown in this paper and in \cref{app:examples} can be experimented with on their original models at \href{http://code2vec.org}{\url{http://code2vec.org}} and \href{https://github.com/microsoft/tf-gnn-samples}{\url{https://github.com/microsoft/tf-gnn-samples}}. 

\Cref{exampleSort} shows additional targeted attacks against the ``sort'' example from \Cref{IndexOfOriginal}. Renaming the variable \scode{array} to \scode{mstyleids} changes the  prediction to \scode{get} with a probability of 99.99\%; renaming \scode{array} to \scode{possiblematches} changes the prediction to \scode{indexOf} with a probability of 86.99\%. The predicted adversarial labels (\scode{get} and \scode{indexOf}) were chosen arbitrarily before finding the variable name replacements.

\para{Transferability}
Occasionally, a dead code attack is \emph{transferable} across examples, and has the same effect even in different examples. This is demonstrated in \Cref{deadcodeExample}, where adding the unused variable declaration \scode{int introsorter = 0;} to each of the multiple snippets changes their prediction to \scode{sort} with probability of $100\%$. 
This effect is reminiscent of the Adversarial Patch \cite{brown2017adversarial}, that was shown to force an image classifier to predict a specific label, regardless of the input example.
However, except for a few cases, we found that adversarial examples \emph{generally do not transfer} across examples. We also did not find significant evidence that adversarial examples transfer across models that were trained on the same dataset, e.g., from \sname{GNN-FiLM} to \sname{GGNN}. The question of whether adversarial examples are transferable in discrete domains such as code remains open. 

\para{GNN Examples}
\Cref{gnnPropertyEmitterShort} shows A C\# \sname{VarMisuse} example that is classified correctly as \scode{\_getterBuilder} in the method \scode{GetGetter} by the \sname{GGNN} model. Given the code and the target \scode{\_setterBuilder}, our approach renames a local variable \scode{setteril} in \emph{another method}. This makes the model predict the wrong variable in the \scode{GetGetter} method, thus introducing a real bug in the method \scode{GetGetter}. \Cref{gnnRequestTypesExample} shows a similar GNN targeted adversarial example.

Other \sname{code2vec} and GNNs examples are shown in \Cref{app:examples}.

%% file: example_sort.tex
\begin{figure*}[t]
\begin{minipage}{\textwidth}
\centering

\begin{tabular}{cc}
\hspace{30mm}
\begin{subfigure}[t]{0.5\textwidth}
\begin{minted}[fontsize=\tiny, frame=single,framesep=2pt,escapeinside=||]{text}
void f(int[] |\boldgreen{array}|){
  boolean swapped = true;
  for (int i = 0; i < |\boldgreen{array}|.length && swapped; i++){
    swapped = false;
    for (int j = 0; j < |\boldgreen{array}|.length-1-i; j++) {
      if (|\boldgreen{array}|[j] > |\boldgreen{array}|[j+1]) {
        int temp = |\boldgreen{array}|[j];
        |\boldgreen{array}|[j] = |\boldgreen{array}|[j+1];
        |\boldgreen{array}|[j+1]= temp;
        swapped = true;
      }
    }
  }
}
\end{minted}
\end{subfigure}
\hspace{-47mm}

\\

\multicolumn{2}{c}{\fbox{\footnotesize Prediction: \boldgreen{\scode{sort} ($98.54\%$)}} }

\\

\begin{subfigure}[t]{0.47\textwidth}
\begin{minted}[fontsize=\tiny, frame=single,framesep=2pt,escapeinside=||]{text}
void f(int[] |\boldmaroon{mstyleids}|){
 boolean swapped = true;
 for (int i = 0; 
      i < |\boldmaroon{mstyleids}|.length && swapped; i++){
  swapped = false;
  for (int j = 0; j < |\boldmaroon{mstyleids}|.length-1-i; j++){
   if (|\boldmaroon{mstyleids}|[j] > |\boldmaroon{mstyleids}|[j+1]) {
    int temp = |\boldmaroon{mstyleids}|[j];
    |\boldmaroon{mstyleids}|[j] = |\boldmaroon{mstyleids}|[j+1];
    |\boldmaroon{mstyleids}|[j+1]= temp;
    swapped = true;
   }
  }
 }
}
\end{minted}
\end{subfigure}

&
\hspace{-3mm}
\begin{subfigure}[t]{0.51\textwidth}
\begin{minted}[fontsize=\tiny, frame=single,framesep=2pt,escapeinside=||]{text}
void f(int[] |\boldmaroon{possiblematches}|){
 boolean swapped = true;
 for (int i = 0; 
      i < |\boldmaroon{possiblematches}|.length && swapped; i++){
  swapped = false;
  for (int j = 0; j < |\boldmaroon{possiblematches}|.length-1-i; j++){
   if (|\boldmaroon{possiblematches}|[j] > |\boldmaroon{possiblematches}|[j+1]){
    int temp = |\boldmaroon{possiblematches}|[j];
    |\boldmaroon{possiblematches}|[j] = |\boldmaroon{possiblematches}|[j+1];
    |\boldmaroon{possiblematches}|[j+1]= temp;
    swapped = true;
   }
  }
 }
}
\end{minted}
\end{subfigure}
\\
\fbox{\footnotesize Prediction: \boldmaroon{\scode{get} ($99.99\%$)}} 
 &
\fbox{\footnotesize Prediction: \boldmaroon{\scode{indexOf} ($86.99\%$)}} \\

\end{tabular}
\end{minipage}
\caption{A snippet classified correctly as \scode{sort} by the model of \href{https://code2vec.org}{\url{code2vec.org}}. The same example is classified as \scode{get} by renaming \scode{array} to \scode{mstyleids} and is classified as \scode{indexOf} by renaming \scode{array} to \scode{possiblematches}.}
\label{exampleSort}
\end{figure*}

%% file: example_deadcode.tex
\begin{figure*}[h!]
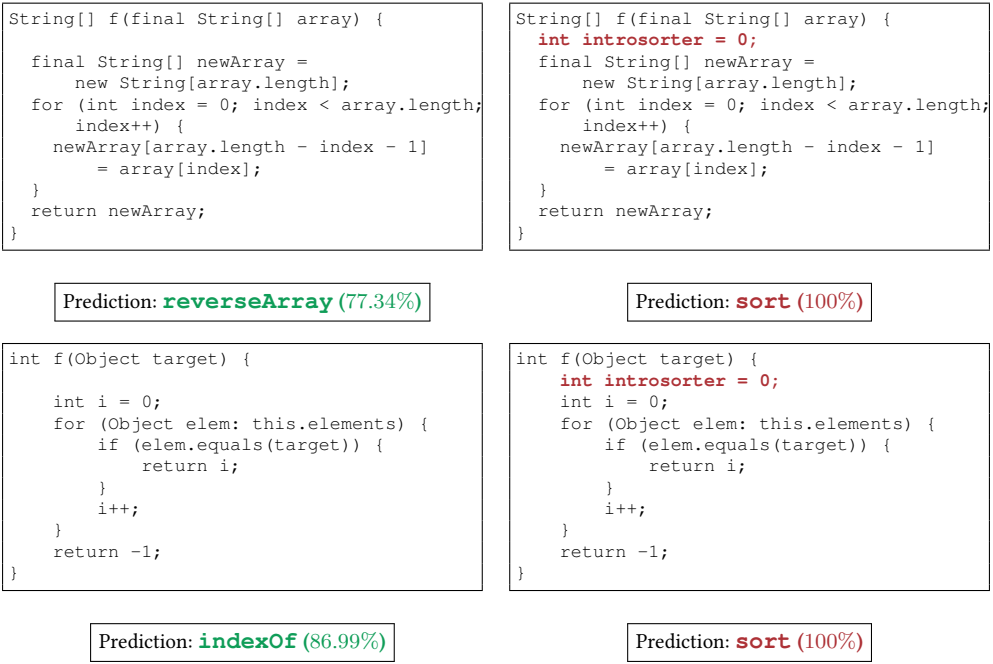

\begin{minipage}{\textwidth}
\centering

\begin{tabular}{ll}

\begin{subfigure}[t]{0.46\textwidth}
\begin{minted}[fontsize=\scriptsize, frame=single,framesep=2pt,escapeinside=||]{text}
String[] f(final String[] array) {

  final String[] newArray = 
      new String[array.length];
  for (int index = 0; index < array.length; 
      index++) {
    newArray[array.length - index - 1]
        = array[index];
  }
  return newArray;
}
\end{minted}
\end{subfigure}

&

\begin{subfigure}[t]{0.46\textwidth}
\begin{minted}[fontsize=\scriptsize, frame=single,framesep=2pt,escapeinside=||]{text}
String[] f(final String[] array) {
  |\boldmaroon{int introsorter = 0;}|
  final String[] newArray = 
      new String[array.length];
  for (int index = 0; index < array.length; 
      index++) {
    newArray[array.length - index - 1] 
        = array[index];
  }
  return newArray;
}
\end{minted}
\end{subfigure}

\\ \\
\multicolumn{1}{c}{
\fbox{\footnotesize Prediction: \boldgreen{\scode{reverseArray} ($77.34\%$)}} 
}
&
\multicolumn{1}{c}{
\fbox{\footnotesize Prediction: \boldmaroon{\scode{sort} ($100\%$)}} 
} \\

\begin{subfigure}[t]{0.46\textwidth}
\begin{minted}[fontsize=\scriptsize, frame=single,framesep=2pt,escapeinside=||]{text}
int f(Object target) {

    int i = 0;
    for (Object elem: this.elements) {
        if (elem.equals(target)) {
            return i;
        }
        i++;
    }
    return -1;
}
\end{minted}
\end{subfigure}
&

\begin{subfigure}[t]{0.46\textwidth}
\begin{minted}[fontsize=\scriptsize, frame=single,framesep=2pt,escapeinside=||]{text}
int f(Object target) {
    |\boldmaroon{int introsorter = 0;}|
    int i = 0;
    for (Object elem: this.elements) {
        if (elem.equals(target)) {
            return i;
        }
        i++;
    }
    return -1;
}
\end{minted}
\end{subfigure}
\\ \\
\multicolumn{1}{c}{
\fbox{\footnotesize Prediction: \boldgreen{\scode{indexOf} ($86.99\%$)}} 
}
&
\multicolumn{1}{c}{ 
\fbox{\footnotesize Prediction: \boldmaroon{\scode{sort} ($100\%$)}}
} \\

\end{tabular}
\end{minipage}
\caption{Adding the dead code \scode{int introsorter = 0;} to each of the snippets on the left changes their label to \scode{sort} with confidence of $100\%$. This is an example of how the same dead adversarial ``patch'' can be applied across different examples.}
\label{deadcodeExample}
\end{figure*}

%% file: example_property_short.tex
\begin{figure*}[t]
\begin{minipage}{\textwidth}
\centering

\begin{tabular}{cc}
\textbf{Correctly predicted example}

\hspace{3mm}
&

\textbf{Adversarial perturbation} \\
& \fbox{\footnotesize Target: \boldmaroon{\texttt{\_setterBuilder}}} \\

\hspace{-2mm}

\begin{subfigure}[t]{0.49\textwidth}
\begin{minted}[fontsize=\fontsize{6}{6}, frame=single,framesep=2pt,escapeinside=**]{text}
public class PropertyEmitter
{
  ...
        
  public *\textbf{PropertyEmitter}*( ... )
    {
    ...
	
    ILGenerator *\boldgreen{setteril}* = 
      _setterBuilder.GetILGenerator();
    *\boldgreen{setteril}*.Emit(OpCodes.Ldarg_0);
    *\boldgreen{setteril}*.Emit(OpCodes.Ldarg_1);
    *\boldgreen{setteril}*.Emit(OpCodes.Stfld, _fieldBuilder);
    if (propertyChangedField != null)
    {
      *\boldgreen{setteril}*.Emit(OpCodes.Ldarg_0);
      *\boldgreen{setteril}*.Emit(OpCodes.Dup);
      *\boldgreen{setteril}*.Emit(
        OpCodes.Ldfld, 
        propertyChangedField);
      *\boldgreen{setteril}*.Emit(OpCodes.Ldstr, name);
      *\boldgreen{setteril}*.Emit(
        OpCodes.Call, 
        ProxyBaseNotifyPropertyChanged);
    }
    *\boldgreen{setteril}*.Emit(OpCodes.Ret);
    _propertyBuilder.SetSetMethod(_setterBuilder);
  }

  ...

  public MethodBuilder *\textbf{GetGetter}*(Type requiredType) 
    => !requiredType.IsAssignableFrom(PropertyType)
    ? throw new InvalidOperationException(
        "Types are not compatible")
    : *\slotbox{\boldgreen{\_getterBuilder}}*;
}       
\end{minted}
\caption{}
\label{gnnSetterBuilderCorrect}
\end{subfigure}
&

\begin{subfigure}[t]{0.49\textwidth}
\begin{minted}[fontsize=\fontsize{6}{6}, frame=single,framesep=2pt,escapeinside=**]{text}
public class PropertyEmitter
{
  ...

  public *\textbf{PropertyEmitter}*( ... )
    {
    ...
    
    ILGenerator *\boldmaroon{va}* = 
      _setterBuilder.GetILGenerator();
    *\boldmaroon{va}*.Emit(OpCodes.Ldarg_0);
    *\boldmaroon{va}*.Emit(OpCodes.Ldarg_1);
    *\boldmaroon{va}*.Emit(OpCodes.Stfld, _fieldBuilder);
    if (propertyChangedField != null)
    {
      *\boldmaroon{va}*.Emit(OpCodes.Ldarg_0);
      *\boldmaroon{va}*.Emit(OpCodes.Dup);
      *\boldmaroon{va}*.Emit(
        OpCodes.Ldfld, 
        propertyChangedField);
      *\boldmaroon{va}*.Emit(OpCodes.Ldstr, name);
      *\boldmaroon{va}*.Emit(
        OpCodes.Call, 
        ProxyBaseNotifyPropertyChanged);
    }
    *\boldmaroon{va}*.Emit(OpCodes.Ret);
    _propertyBuilder.SetSetMethod(_setterBuilder);
  }

  ...

  public MethodBuilder *\textbf{GetGetter}*(Type requiredType) 
    => !requiredType.IsAssignableFrom(PropertyType)
    ? throw new InvalidOperationException(
        "Types are not compatible")
    : *\slotbox{\boldmaroon{\_setterBuilder}}*;
}       
\end{minted}
\caption{}
\label{gnnSetterBuilderIncorrect}
\end{subfigure}

\end{tabular}
\end{minipage}
\caption{A C\# \sname{VarMisuse} example which is classified correctly as \scode{\_getterBuilder} in the method \scode{GetGetter} by the \sname{GGNN} model. Given the code and the target \scode{\_setterBuilder}, our approach renames a local variable \scode{setteril} in \emph{another method}. This makes the model predict the wrong variable in the \scode{GetGetter} method, thus introducing a real bug in the method \scode{GetGetter}. }
\label{gnnPropertyEmitterShort}
\end{figure*}

%% file: example_gnn_destinationType.tex
\begin{figure*}[t]
\begin{minipage}{\textwidth}
\centering

\begin{tabular}{cc}
\textbf{Correctly predicted example}

\hspace{3mm}
&

\textbf{Adversarial perturbation} \\
& \fbox{\footnotesize Target: \boldmaroon{\texttt{RequestedTypes}}} \\

\hspace{-2mm}

\begin{subfigure}[t]{0.47\textwidth}
\begin{minted}[fontsize=\scriptsize, frame=single,framesep=2pt,escapeinside=||]{text}
public MapRequest(TypePair |\boldgreen{requestedTypes}|, 
    TypePair runtimeTypes, 
    IMemberMap memberMap = null) 
{
  RequestedTypes = |\boldgreen{requestedTypes}|;
  |\slotbox{\boldgreen{RuntimeTypes}}| = runtimeTypes;
  MemberMap = memberMap;
}
\end{minted}
\caption{}
\label{gnnRequestedTypesCorrect}
\end{subfigure}
&
\hspace{-4mm}

\begin{subfigure}[t]{0.51\textwidth}
\begin{minted}[fontsize=\scriptsize, frame=single,framesep=2pt,escapeinside=||]{text}
public MapRequest(TypePair |\boldmaroon{chzuzb}|, 
    TypePair runtimeTypes, 
    IMemberMap memberMap = null) 
{
  RequestedTypes = |\boldmaroon{chzuzb}|;
  |\slotbox{\boldmaroon{RequestedTypes}}| = runtimeTypes;
  MemberMap = memberMap;
}
\end{minted}
\caption{}
\label{gnnRequestedTypesIncorrect}
\end{subfigure}

\end{tabular}
\end{minipage}
\caption{A C\# \sname{VarMisuse} example that is classified correctly by the \sname{GGNN} model. Given the code and the target \scode{RequestedTypes}, our approach renames a local variable \scode{requestedTypes}. This makes the model predict the wrong variable, thus introducing a real bug. }
\label{gnnRequestTypesExample}
\end{figure*}

%% file: related.tex
\section{Related Work}

\para{Adversarial Examples of Images} \citet{szegedy2013intriguing} found that deep neural networks are vulnerable to adversarial examples. They showed that 
they could cause an image classification model to misclassify an image by applying a certain barely perceptible perturbation. \citet{goodfellow2014explaining} introduced an efficient method called ``fast gradient sign method'' to generate adversarial examples. Their method was based on adding an imperceptibly small vector whose elements are equal to the \emph{sign} of the elements of the gradient of the cost function with respect to the input. 
Generating adversarial examples in images is probably easier than in discrete domains such as code and natural language. Images are continuous objects, and thus can be perturbed with small undetected noise; in contrast, our problem domain is \emph{discrete}.

\para{Adversarial Examples in NLP}
The challenge of adversarial examples for \emph{discrete inputs} has been studied in the domain of NLP. 
While adversarial examples on images are easy to generate, the generation of adversarial text is more difficult. 
Images are continuous and can thus have some noise added; natural language text is discrete and thus cannot be easily perturbed. 
HotFlip \cite{ebrahimi2017hotflip} is a technique for generating adversarial examples that attack a character-level neural classifier. The idea is to flip a single character in a word, producing a typo such that the change is hardly noticeable. %
\citet{alzantot2018generating} presented a technique for generating ``semantically and syntactically similar adversarial examples'' that attack trained models. The main idea is to replace a random word in a given sentence with a similar word (nearest neighbor) in some embedding space. 
However, these techniques allow only \emph{non-targeted} attacks, i.e., impair a correct prediction, without aiming for a specific adversarial label.

\para{Adversarial Examples in Malware}
A few works explored adversarial examples in malware detection that have the ability to perturb a malicious binary such that a learning model will classify it as ``benign''. %
\citet{kreuk2018deceiving}, \citet{suciu2019exploring} and \citet{kolosnjaji2018adversarial} addressed binary classifiers (whether or not the program is malicious) of binary code, 
by adding noisy bytes to the original file's raw bytes. None of these works performed \emph{targeted} attacks as does our work.
In most cases, hiding an adversarial dead-code payload inside a binary may be easier than generating adversarial examples in high-level languages like Java or C\#. For example, \citet{kolosnjaji2018adversarial} reported that they injected at least 10,000 ``padding bytes'' to each malware sample; because a binary file is usually much larger, injecting 10,000 bytes can go unnoticed. In contrast, in all our attacks we renamed a \emph{single} variable or added a \emph{single} variable declaration.
Another difference from our approach is that all these works derive the loss based on the embedding vector itself, as we discuss in \Cref{subsec:method}. In contrast, we derive the loss by the distribution over indices (\Cref{subsec:deriving}), which directly optimizes towards the target label; this allows us to perform targeted attacks in multi-class models. 
\citet{rosenberg2018generic} addressed a very different scenario and modified malicious programs to mimic benign calls to APIs \emph{at runtime}. \citet{yang2017malware} presented a black-box approach to attack non-neural models. Their approach is not gradient-based and thus cannot perform targeted attacks.

\para{Adversarial Examples in High-Level Programs}
To the best of our knowledge, our work is the first to investigate adversarial attacks for models of high-level code. \citet{rabintesting} identified a problem of robustness in models of code, but did not suggest a concrete method for producing adversarial examples or a method for defending against them.

\para{Defending Against Adversarial Examples} 
A few concurrent work with ours addressed the problem of training models of code to be more robust. \citet{ramakrishnan2020semantic} performed semantic-preserving transformations such as renaming, and trained neural models on the modified code. 
However, their transformations are applied in the data preprocessing step, and their approach does not use gradients or targeted attacks.
\citet{bielik2020adversarial} focused on training robust models as well, with an iterative approach for adversarial training. These works only discussed approaches for adversarial training, but none presented \emph{targeted attacks}, as our work.

\citet{pruthi2019combating} found that NLP models perform much worse when the input contains spelling mistakes. To make these models more robust to misspellings, the authors placed a character-level word recognition model in front of the downstream model. This word recognition model repairs misspellings before they are fed into the downstream model. 
The \emph{Outlier Detection} defense that we examined (\Cref{subsec:without-retraining}) is similar in spirit, since it uses a composition of an upstream defense model followed by a downstream model. 
The main difference between these two defenses resides in the goals of the upstream defense models: the \emph{Outlier Detection} model detects outlier names, while the model of \citet{pruthi2019combating} detects character-level typos.

%% file: conclusion.tex
\section{Conclusion}
We presented \method{}, the first approach for generating \emph{targeted} attacks on models of code using adversarial examples. Our approach is a \emph{general white-box technique that can work for any model of code in which we can compute gradients.} 
We demonstrated \method{} on popular neural architectures for code: \sname{code2vec}, \sname{GGNN}, and \sname{GNN-FiLM}, in Java and C\#.
Moreover, we showed that \method{} succeeds in both targeted and non-targeted attacks, by renaming variables and by adding dead code.

We further experimented with a variety of possible defense techniques and discuss their trade-offs across performance, robustness, and whether or not they require re-training. 

We believe the principles presented in this paper can be the basis for a wide range of adversarial attacks and defenses. %
This is the first work to perform targeted attacks for models of code; thus, using our attack in adversarial training contributes to more robust models. 
In realistic production environments, the defense techniques that we examine can further strengthen robustness. 
 To this end, we make all our code, data, and trained models publicly available at \url{https://github.com/tech-srl/adversarial-examples} .

%% file: ack.tex
\section*{Acknowledgements}
We would like to thank Marc Brockschmidt and Miltiadis Allamanis for their guidance in using their GNN models and code and useful discussions about the differences in robustness between GNN types. 

The research leading to these results has received funding from
the Israel Ministry of Science and Technology, grant no. 3-9779.

%% file: example_escape.tex
\begin{figure*}[h!]
\begin{minipage}{\textwidth}
\centering

\begin{tabular}{cc}
\hspace{36mm}
\begin{subfigure}[t]{0.45\textwidth}
\begin{minted}[fontsize=\scriptsize, frame=single,framesep=2pt,escapeinside=||]{text}
String f(String |\boldgreen{txt}|) {
 |\boldgreen{txt}| = replace(|\boldgreen{txt}|, "&", "&amp;");
 |\boldgreen{txt}| = replace(|\boldgreen{txt}|, "\"", "&quote;");
 |\boldgreen{txt}| = replace(|\boldgreen{txt}|, "<", "&lt;");
 |\boldgreen{txt}| = replace(|\boldgreen{txt}|, ">", "&gt;");
 return |\boldgreen{txt}|;
}
\end{minted}
\end{subfigure}
\hspace{-36mm}
\\
\multicolumn{2}{c}{\fbox{\footnotesize Prediction: \boldgreen{\scode{escape}}}} \\

\\
\fbox{\footnotesize Target: \boldmaroon{\scode{contains}}} 
&
\fbox{\footnotesize Target: \boldmaroon{\scode{done}}}

\\
\begin{subfigure}[t]{0.47\textwidth}
\begin{minted}[fontsize=\scriptsize, frame=single,framesep=2pt,escapeinside=||]{text}
String f(String |\boldmaroon{expres}|) {
 |\boldmaroon{expres}| = replace(|\boldmaroon{expres}|, "&", "&amp;");
 |\boldmaroon{expres}| = replace(|\boldmaroon{expres}|, "\"", "&quote;");
 |\boldmaroon{expres}| = replace(|\boldmaroon{expres}|, "<", "&lt;");
 |\boldmaroon{expres}| = replace(|\boldmaroon{expres}|, ">", "&gt;");
 return |\boldmaroon{expres}|;
}
\end{minted}
\end{subfigure}

&

\begin{subfigure}[t]{0.49\textwidth}
\begin{minted}[fontsize=\scriptsize, frame=single,framesep=2pt,escapeinside=||]{text}
String f(String |\boldmaroon{claimed}|) {
 |\boldmaroon{claimed}| = replace(|\boldmaroon{claimed}|, "&", "&amp;");
 |\boldmaroon{claimed}| = replace(|\boldmaroon{claimed}|, "\"", "&quote;");
 |\boldmaroon{claimed}| = replace(|\boldmaroon{claimed}|, "<", "&lt;");
 |\boldmaroon{claimed}| = replace(|\boldmaroon{claimed}|, ">", "&gt;");
 return |\boldmaroon{claimed}|;
}
\end{minted}
\end{subfigure}
\\

\begin{subfigure}[t]{0.47\textwidth}
\centering
\fbox{\footnotesize Prediction: \boldmaroon{\scode{contains} ($94.99\%$)}} \\
\end{subfigure}

&
\begin{subfigure}[t]{0.49\textwidth}
\centering
\fbox{\footnotesize Prediction: \boldmaroon{\scode{done} ($77.68\%$)}} \\
\end{subfigure}

\end{tabular}
\end{minipage}
\caption{A snippet classified correctly as \scode{escape} by the model of \href{https://code2vec.org}{\url{code2vec.org}}. The same example is classified as \scode{contains} by renaming \scode{txt} to \scode{expres} and is classified as \scode{done} by renaming \scode{txt} to \scode{claimed}. These targets (\scode{contains} and \scode{done}) were chosen arbitrarily in advance, and our \method{} method had found the new required variable names \scode{expres} and \scode{claimed}.}
\label{exampleContains}
\end{figure*}

%% file: example_contains.tex
\begin{figure*}[h!]
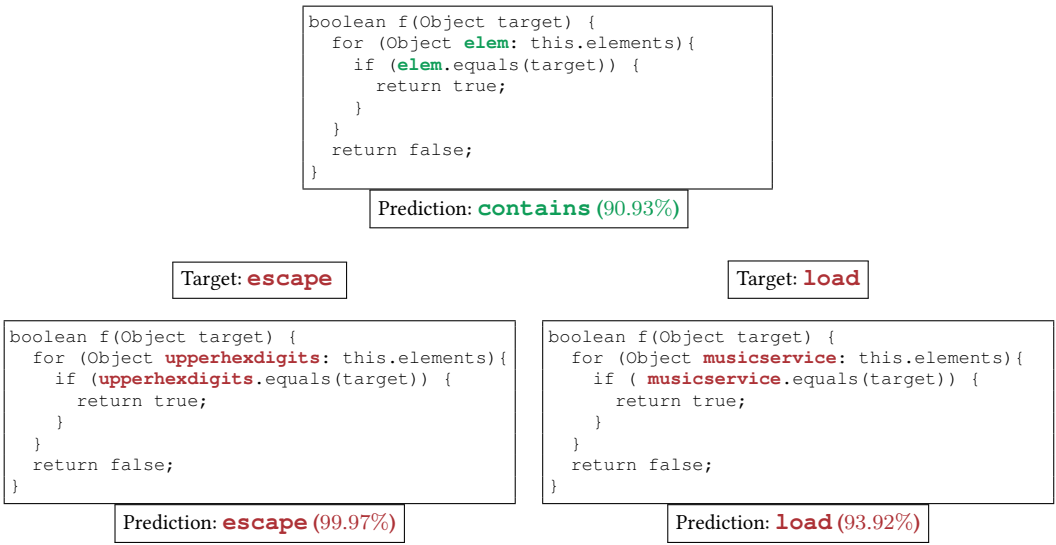

\begin{minipage}{\textwidth}
\centering
\begin{tabular}{cc}
\hspace{37mm}
\begin{subfigure}[t]{0.45\textwidth}
\begin{minted}[fontsize=\scriptsize, frame=single,framesep=2pt,escapeinside=||]{text}
boolean f(Object target) {
  for (Object |\boldgreen{elem}|: this.elements){
    if (|\boldgreen{elem}|.equals(target)) {
      return true;
    }
  }
  return false;
}
\end{minted}
\end{subfigure}
\hspace{-37mm}

\\
\multicolumn{2}{c}{\fbox{\footnotesize Prediction: \boldgreen{\scode{contains} ($90.93\%$)}}}

\\
\\
\fbox{\footnotesize Target: \boldmaroon{\scode{escape} }}
&
\fbox{\footnotesize Target: \boldmaroon{\scode{load}}} \\

\begin{subfigure}[t]{0.49\textwidth}
\begin{minted}[fontsize=\scriptsize, frame=single,framesep=2pt,escapeinside=||]{text}
boolean f(Object target) {
  for (Object |\boldmaroon{upperhexdigits}|: this.elements){
    if (|\boldmaroon{upperhexdigits}|.equals(target)) {
      return true;
    }
  }
  return false;
}
\end{minted}
\end{subfigure}

&
\begin{subfigure}[t]{0.49\textwidth}
\begin{minted}[fontsize=\scriptsize, frame=single,framesep=2pt,escapeinside=||]{text}
boolean f(Object target) {
  for (Object |\boldmaroon{musicservice}|: this.elements){
    if ( |\boldmaroon{musicservice}|.equals(target)) {
      return true;
    }
  }
  return false;
}
\end{minted}
\end{subfigure}
\\
\fbox{\footnotesize Prediction: \boldmaroon{\scode{escape} ($99.97\%$)}}
&
\fbox{\footnotesize Prediction: \boldmaroon{\scode{load} ($93.92\%$)}} \\

\end{tabular}
\end{minipage}
\caption{A snippet classified correctly as \scode{contains} by the model of \href{https://code2vec.org}{\url{code2vec.org}}. The same example is classified as \scode{escape} by renaming \scode{elem} to \scode{upperhexdigits} and is classified as \scode{load} by renaming \scode{elem} to \scode{musicservice}. These targets (\scode{escape} and \scode{load}) were chosen arbitrarily in advance, and our \method{} method had found the new required variable names \scode{upperhexdigits} and \scode{musicservice}.}
\label{exampleContains}
\end{figure*}

%% file: example_count.tex
\begin{figure*}[h!]
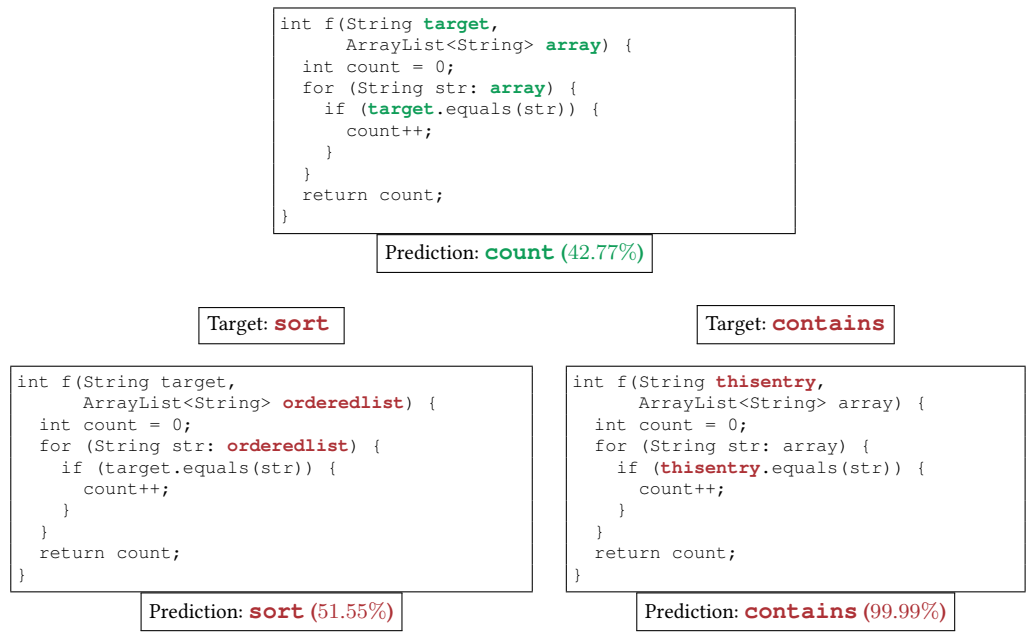

\begin{minipage}{\textwidth}
\centering

\begin{tabular}{cc}
\hspace{35mm}
\begin{subfigure}[t]{0.5\textwidth}
\begin{minted}[fontsize=\scriptsize, frame=single,framesep=2pt,escapeinside=||]{text}
int f(String |\boldgreen{target}|, 
      ArrayList<String> |\boldgreen{array}|) {
  int count = 0;
  for (String str: |\boldgreen{array}|) {
    if (|\boldgreen{target}|.equals(str)) {
      count++;
    }
  }
  return count;
}
\end{minted}
\end{subfigure}
\hspace{-35mm}

\\
\multicolumn{2}{c}{\fbox{\footnotesize Prediction: \boldgreen{\scode{count} ($42.77\%$)}}}

\\ \\
\fbox{\footnotesize Target: \boldmaroon{\scode{sort} }}
&

\fbox{\footnotesize Target: \boldmaroon{\scode{contains}}} \\
\begin{subfigure}[t]{0.5\textwidth}
\begin{minted}[fontsize=\scriptsize, frame=single,framesep=2pt,escapeinside=||]{text}
int f(String target, 
      ArrayList<String> |\boldmaroon{orderedlist}|) {
  int count = 0;
  for (String str: |\boldmaroon{orderedlist}|) {
    if (target.equals(str)) {
      count++;
    }
  }
  return count;
}
\end{minted}
\end{subfigure}

& 

\begin{subfigure}[t]{0.44\textwidth}
\begin{minted}[fontsize=\scriptsize, frame=single,framesep=2pt,escapeinside=||]{text}
int f(String |\boldmaroon{thisentry}|, 
      ArrayList<String> array) {
  int count = 0;
  for (String str: array) {
    if (|\boldmaroon{thisentry}|.equals(str)) {
      count++;
    }
  }
  return count;
}
\end{minted}
\end{subfigure}  

\\ 
\fbox{\footnotesize Prediction: \boldmaroon{\scode{sort} ($51.55\%$)}}
&

\fbox{\footnotesize Prediction: \boldmaroon{\scode{contains} ($99.99\%$)}} \\

\end{tabular}
\end{minipage}
\caption{A snippet classified correctly as \scode{count} by the model of \href{https://code2vec.org}{\url{code2vec.org}}. The same example is classified as \scode{sort} by renaming \scode{array} to \scode{orderedlist} and is classified as \scode{contains} by renaming \scode{target} to \scode{thisentry}. These targets (\scode{sort} and \scode{contains}) were chosen arbitrarily in advance, and our \method{} method had found the new required variable names \scode{orderedlist} and \scode{thisentry}.}
\label{exampleContains}
\end{figure*}

%% file: example_propertyEmitter.tex
\begin{figure*}[t]
\begin{minipage}{\textwidth}
\centering

\begin{tabular}{cc}
\textbf{Correctly predicted example}

\hspace{3mm}
&

\textbf{Adversarial perturbation} \\
& \fbox{\footnotesize Target: \boldmaroon{\texttt{\_setterBuilder}}} \\

\hspace{-2mm}

\begin{subfigure}[t]{0.49\textwidth}
\begin{minted}[fontsize=\fontsize{5.8}{5.8}, frame=single,framesep=2pt,escapeinside=**]{text}
public class PropertyEmitter
{
  ...
        
  private readonly FieldBuilder _fieldBuilder;
  private readonly MethodBuilder _getterBuilder;
  private readonly PropertyBuilder _propertyBuilder;
  private readonly MethodBuilder _setterBuilder;

  public *\textbf{PropertyEmitter}*(
      TypeBuilder owner, 
      PropertyDescription property, 
      FieldBuilder propertyChangedField)
    {
    var name = property.Name;
    var propertyType = property.Type;
    _fieldBuilder = owner.DefineField($"<{name}>", 
      propertyType, FieldAttributes.Private);
    _propertyBuilder = owner.DefineProperty(name, 
      PropertyAttributes.None, propertyType, null);
    _getterBuilder = owner.DefineMethod($"get_{name}",
      MethodAttributes.Public 
        | MethodAttributes.Virtual 
        | MethodAttributes.HideBySig 
        | MethodAttributes.SpecialName, 
      propertyType, 
      Type.EmptyTypes);
    ILGenerator getterIl = 
      _getterBuilder.GetILGenerator();
    getterIl.Emit(OpCodes.Ldarg_0);
    getterIl.Emit(OpCodes.Ldfld, _fieldBuilder);
    getterIl.Emit(OpCodes.Ret);
    _propertyBuilder.SetGetMethod(_getterBuilder);
    if(!property.CanWrite)
    {
      return;
    }
    _setterBuilder = owner.DefineMethod($"set_{name}",
      MethodAttributes.Public 
        | MethodAttributes.Virtual 
        | MethodAttributes.HideBySig 
        | MethodAttributes.SpecialName, 
      typeof (void), 
      new[] {propertyType});
    ILGenerator *\boldgreen{setteril}* = 
      _setterBuilder.GetILGenerator();
    *\boldgreen{setteril}*.Emit(OpCodes.Ldarg_0);
    *\boldgreen{setteril}*.Emit(OpCodes.Ldarg_1);
    *\boldgreen{setteril}*.Emit(OpCodes.Stfld, _fieldBuilder);
    if (propertyChangedField != null)
    {
      *\boldgreen{setteril}*.Emit(OpCodes.Ldarg_0);
      *\boldgreen{setteril}*.Emit(OpCodes.Dup);
      *\boldgreen{setteril}*.Emit(
        OpCodes.Ldfld, 
        propertyChangedField);
      *\boldgreen{setteril}*.Emit(OpCodes.Ldstr, name);
      *\boldgreen{setteril}*.Emit(
        OpCodes.Call, 
        ProxyBaseNotifyPropertyChanged);
    }
    *\boldgreen{setteril}*.Emit(OpCodes.Ret);
    _propertyBuilder.SetSetMethod(_setterBuilder);
  }

  ...

  public MethodBuilder *\textbf{GetGetter}*(Type requiredType) 
    => !requiredType.IsAssignableFrom(PropertyType)
    ? throw new InvalidOperationException(
        "Types are not compatible")
    : *\slotbox{\boldgreen{\_getterBuilder}}*;
}       
\end{minted}
\caption{}
\label{gnnSetterBuilderCorrect}
\end{subfigure}
&

\begin{subfigure}[t]{0.49\textwidth}
\begin{minted}[fontsize=\fontsize{5.8}{5.8}, frame=single,framesep=2pt,escapeinside=**]{text}
public class PropertyEmitter
{
  ...
        
  private readonly FieldBuilder _fieldBuilder;
  private readonly MethodBuilder _getterBuilder;
  private readonly PropertyBuilder _propertyBuilder;
  private readonly MethodBuilder _setterBuilder;

  public *\textbf{PropertyEmitter}*(
      TypeBuilder owner, 
      PropertyDescription property, 
      FieldBuilder propertyChangedField)
    {
    var name = property.Name;
    var propertyType = property.Type;
    _fieldBuilder = owner.DefineField($"<{name}>", 
      propertyType, FieldAttributes.Private);
    _propertyBuilder = owner.DefineProperty(name, 
      PropertyAttributes.None, propertyType, null);
    _getterBuilder = owner.DefineMethod($"get_{name}",
      MethodAttributes.Public 
        | MethodAttributes.Virtual 
        | MethodAttributes.HideBySig 
        | MethodAttributes.SpecialName, 
      propertyType, 
      Type.EmptyTypes);
    ILGenerator getterIl = 
      _getterBuilder.GetILGenerator();
    getterIl.Emit(OpCodes.Ldarg_0);
    getterIl.Emit(OpCodes.Ldfld, _fieldBuilder);
    getterIl.Emit(OpCodes.Ret);
    _propertyBuilder.SetGetMethod(_getterBuilder);
    if(!property.CanWrite)
    {
      return;
    }
    _setterBuilder = owner.DefineMethod($"set_{name}",
      MethodAttributes.Public 
        | MethodAttributes.Virtual 
        | MethodAttributes.HideBySig 
        | MethodAttributes.SpecialName, 
      typeof (void), 
      new[] {propertyType});
    ILGenerator *\boldmaroon{va}* = 
      _setterBuilder.GetILGenerator();
    *\boldmaroon{va}*.Emit(OpCodes.Ldarg_0);
    *\boldmaroon{va}*.Emit(OpCodes.Ldarg_1);
    *\boldmaroon{va}*.Emit(OpCodes.Stfld, _fieldBuilder);
    if (propertyChangedField != null)
    {
      *\boldmaroon{va}*.Emit(OpCodes.Ldarg_0);
      *\boldmaroon{va}*.Emit(OpCodes.Dup);
      *\boldmaroon{va}*.Emit(
        OpCodes.Ldfld, 
        propertyChangedField);
      *\boldmaroon{va}*.Emit(OpCodes.Ldstr, name);
      *\boldmaroon{va}*.Emit(
        OpCodes.Call, 
        ProxyBaseNotifyPropertyChanged);
    }
    *\boldmaroon{va}*.Emit(OpCodes.Ret);
    _propertyBuilder.SetSetMethod(_setterBuilder);
  }

  ...

  public MethodBuilder *\textbf{GetGetter}*(Type requiredType) 
    => !requiredType.IsAssignableFrom(PropertyType)
    ? throw new InvalidOperationException(
        "Types are not compatible")
    : *\slotbox{\boldmaroon{\_setterBuilder}}*;
}       
\end{minted}
\caption{}
\label{gnnSetterBuilderIncorrect}
\end{subfigure}

\end{tabular}
\end{minipage}
\caption{A C\# \sname{VarMisuse} example which is classified correctly as \scode{\_getterBuilder} in the method \scode{GetGetter} by the \sname{GGNN} model. Given the code and the target \scode{\_setterBuilder}, our approach renames a local variable \scode{setteril} in \emph{another method}, making the model predict the wrong variable in the \scode{GetGetter} method, thus introducing a real bug in the method \scode{GetGetter}.}
\label{gnnPropertyEmitter}
\end{figure*}

%% file: example_proxyGenerator.tex
\begin{figure*}[t]
\begin{minipage}{\textwidth}
\centering

\begin{tabular}{cc}
\textbf{Correctly predicted example}

\hspace{3mm}
&

\textbf{Adversarial perturbation} \\
& \fbox{\footnotesize Target: \boldmaroon{\texttt{typeBuilder}}} \\

\hspace{-2mm}

\begin{subfigure}[t]{0.49\textwidth}
\begin{minted}[fontsize=\fontsize{6}{6}, frame=single,framesep=2pt,escapeinside=**]{text}
public static class ProxyGenerator
{
    ...

  private static *\textbf{ModuleBuilder}* CreateProxyModule()
  {
    AssemblyName name = new AssemblyName(
      "AutoMapper.Proxies");
    name.SetPublicKey(privateKey);
    name.SetPublicKeyToken(privateKeyToken);

    AssemblyBuilder *\boldgreen{builder}* = 
      AssemblyBuilder.DefineDynamicAssembly(
        name, AssemblyBuilderAccess.Run);

    return *\boldgreen{builder}*.DefineDynamicModule(
      "AutoMapper.Proxies.emit");
  }

  private static Type *\textbf{EmitProxy}*(
    TypeDescription typeDescription)
  {
    var interfaceType = typeDescription.Type;
    
    ...
    
    if(typeof(INotifyPropertyChanged)
        .IsAssignableFrom(interfaceType))
    {
      ...
    }
    
    ...

    foreach(var property in propertiesToImplement)
    {
      if(fieldBuilders.TryGetValue(property.Name, 
        out var propertyEmitter))
      {
        if((propertyEmitter.PropertyType 
          != property.Type) && (
            (property.CanWrite) 
            || (!property.Type.IsAssignableFrom(
                   propertyEmitter.PropertyType))))
        {
          throw new ArgumentException(
            $"The interface has a conflicting 
              property {property.Name}",
            nameof(*\slotbox{\boldgreen{interfaceType}}*));
        }
      }
      else
      {
        fieldBuilders.Add(property.Name,
          new PropertyEmitter(
            typeBuilder, 
            property, propertyChangedField));
      }
    }
    return typeBuilder.CreateType();
  }
}     
\end{minted}
\caption{}
\label{gnnTypeBuilderCorrect}
\end{subfigure}
&
\hspace{-4mm}

\begin{subfigure}[t]{0.49\textwidth}
\begin{minted}[fontsize=\fontsize{6}{6}, frame=single,framesep=2pt,escapeinside=**]{text}
public static class ProxyGenerator
{
    ...

  private static *\textbf{ModuleBuilder}* CreateProxyModule()
  {
    AssemblyName name = new AssemblyName(
      "AutoMapper.Proxies");
    name.SetPublicKey(privateKey);
    name.SetPublicKeyToken(privateKeyToken);

    AssemblyBuilder *\boldmaroon{bzodhi}* = 
      AssemblyBuilder.DefineDynamicAssembly(
        name, AssemblyBuilderAccess.Run);

    return *\boldmaroon{bzodhi}*.DefineDynamicModule(
      "AutoMapper.Proxies.emit");
  }

  private static Type *\textbf{EmitProxy}*(
    TypeDescription typeDescription)
  {
    var interfaceType = typeDescription.Type;
    
    ...
    
    if(typeof(INotifyPropertyChanged)
        .IsAssignableFrom(interfaceType))
    {
      ...
    }
    
    ...

    foreach(var property in propertiesToImplement)
    {
      if(fieldBuilders.TryGetValue(property.Name, 
        out var propertyEmitter))
      {
        if((propertyEmitter.PropertyType 
          != property.Type) && (
            (property.CanWrite) 
            || (!property.Type.IsAssignableFrom(
                   propertyEmitter.PropertyType))))
        {
          throw new ArgumentException(
            $"The interface has a conflicting 
              property {property.Name}",
            nameof(*\slotbox{\boldmaroon{typeBuilder}}*));
        }
      }
      else
      {
        fieldBuilders.Add(property.Name,
          new PropertyEmitter(
            typeBuilder, 
            property, propertyChangedField));
      }
    }
    return typeBuilder.CreateType();
  }
}
\end{minted}
\caption{}
\label{gnnTypeBuilderIncorrect}
\end{subfigure}

\end{tabular}
\end{minipage}
\caption{A C\# \sname{VarMisuse} example which is classified correctly as \scode{interfaceType} in the method \scode{EmitProxy} by the \sname{GGNN} model. Given the code and the target \scode{typeBuilder}, our approach renames a local variable \scode{builder} in \emph{another method}, making the model predict the wrong variable in the \scode{EmitProxy} method, thus introducing a real bug in the method \scode{EmitProxy}.}
\label{gnnDestinationType}
\end{figure*}

%% file: example_gnn_SourceType2.tex
\begin{figure*}[t]
\begin{minipage}{\textwidth}
\centering

\begin{tabular}{cc}
\textbf{Correctly predicted example}

\hspace{3mm}
&

\textbf{Adversarial perturbation}\\
&  \fbox{\footnotesize Target: \boldmaroon{\texttt{DestinationType}}} \\

\hspace{-2mm}

\begin{subfigure}[t]{0.49\textwidth}
\begin{minted}[fontsize=\fontsize{6.2}{6.2}, frame=single,framesep=2pt,escapeinside=||]{text}
struct TypePair : IEquatable<TypePair>
{
  public static TypePair 
      |\textbf{Create}|<TSource, TDestination>(
    TSource |\boldgreen{source}|, TDestination destination, 
    Type sourceType, Type destinationType)
  {
    if(|\boldgreen{source}| != null)
    {
      sourceType = |\boldgreen{source}|.GetType();
    }
    if(destination != null)
    {
      destinationType = destination.GetType();
    }
      return new TypePair(
        sourceType, destinationType);
  }
  
  ...
  
  public Type SourceType { get; }
  public Type DestinationType { get; }
  
  ...
  
  public TypePair? |\textbf{GetOpenGenericTypePair}|()
  {
    if(!IsGeneric)
    {
      return null;
    }
    var sourceGenericDefinition = 
      SourceType.IsGenericType() ?
      |\slotbox{\boldgreen{SourceType}}|.GetGenericTypeDefinition() 
          : SourceType;
    var destinationGenericDefinition = 
      DestinationType.IsGenericType() ? 
      DestinationType.GetGenericTypeDefinition() : 
        DestinationType;

    return new TypePair(sourceGenericDefinition, 
      destinationGenericDefinition);
  }
} 
\end{minted}
\caption{}
\label{gnnDestinationType2Correct}
\end{subfigure}
&
\hspace{-4mm}

\begin{subfigure}[t]{0.50\textwidth}
\begin{minted}[fontsize=\fontsize{6.2}{6.2}, frame=single,framesep=2pt,escapeinside=||]{text}
struct TypePair : IEquatable<TypePair>
{
  public static TypePair 
      |\textbf{Create}|<TSource, TDestination>(
    TSource |\boldmaroon{ebwhajqa}|, TDestination destination, 
    Type sourceType, Type destinationType)
  {
    if(|\boldmaroon{ebwhajqa}| != null)
    {
      sourceType = |\boldmaroon{ebwhajqa}|.GetType();
    }
    if(destination != null)
    {
      destinationType = destination.GetType();
    }
      return new TypePair(
        sourceType, destinationType);
  }
  
  ...
  
  public Type SourceType { get; }
  public Type DestinationType { get; }
  
  ...
  
  public TypePair? |\textbf{GetOpenGenericTypePair}|()
  {
    if(!IsGeneric)
    {
      return null;
    }
    var sourceGenericDefinition = 
      SourceType.IsGenericType() ?
      |\slotbox{\boldmaroon{DestinationType}}|.GetGenericTypeDefinition() 
          : SourceType;
    var destinationGenericDefinition = 
      DestinationType.IsGenericType() ? 
      DestinationType.GetGenericTypeDefinition() : 
        DestinationType;

    return new TypePair(sourceGenericDefinition, 
      destinationGenericDefinition);
  }
} 
\end{minted}
\caption{}
\label{gnnDestinationType2Incorrect}
\end{subfigure}

\end{tabular}
\end{minipage}
\caption{A C\# \sname{VarMisuse} example which is classified correctly by the \sname{GGNN} model. Given the code and the target \scode{DestinationType}, our approach renames a local variable \scode{source} \emph{in another method}, making the model predict the wrong variable, thus introducing a real bug. }
\label{gnnDestinationType}
\end{figure*}

%% file: example_gnn_enum_to_enum.tex
\begin{figure*}[t]
\begin{minipage}{\textwidth}
\centering

\begin{tabular}{cc}
\textbf{Correctly predicted example}

\hspace{3mm}
&

\textbf{Adversarial perturbation} \\ 
& \fbox{\footnotesize Target: \boldmaroon{\texttt{destEnumType}}} \\

\hspace{-2mm}

\begin{subfigure}[t]{0.49\textwidth}
\begin{minted}[fontsize=\fontsize{6.0}{6.0}, frame=single,framesep=2pt,escapeinside=||]{text}
public class EnumToEnumMapper : IObjectMapper
{
  public static TDestination 
      |\textbf{Map}|<TSource, TDestination>(TSource source)
  {
    var sourceEnumType = 
      ElementTypeHelper.GetEnumerationType(
        typeof(TSource));
    var destEnumType = 
      ElementTypeHelper.GetEnumerationType(
        typeof(TDestination));

    if (!Enum.IsDefined(sourceEnumType, source))
    {
      return (TDestination)Enum.ToObject(
        destEnumType, source);
    }

    if (!Enum.GetNames(destEnumType).Contains(
      source.ToString(),
      StringComparer.OrdinalIgnoreCase))
    {
      var underlyingSourceType = 
        Enum.GetUnderlyingType(|\slotbox{\boldgreen{sourceEnumType}}|); 
      var underlyingSourceValue = 
        System.Convert.ChangeType(
          source, underlyingSourceType);

      return (TDestination)Enum.ToObject(
        destEnumType, underlyingSourceValue);
    }

    return (TDestination)Enum.Parse(
      destEnumType, Enum.GetName(
        sourceEnumType, source), true);
  }

  private static readonly MethodInfo MapMethodInfo = 
    typeof(EnumToEnumMapper).GetAllMethods().First(
      _ => _.IsStatic);

  public bool |\textbf{IsMatch}|(TypePair |\boldgreen{context}|)
  {
    var sourceEnumType = 
      ElementTypeHelper.GetEnumerationType(
        |\boldgreen{context}|.SourceType);
    var destEnumType = 
      ElementTypeHelper.GetEnumerationType(
        |\boldgreen{context}|.DestinationType);
    return sourceEnumType != null 
      && destEnumType != null;
  }
}
\end{minted}
\caption{}
\label{gnndestEnumTypeCorrect}
\end{subfigure}
&
\hspace{-4mm}

\begin{subfigure}[t]{0.49\textwidth}
\begin{minted}[fontsize=\fontsize{6.0}{6.0}, frame=single,framesep=2pt,escapeinside=||]{text}
public class EnumToEnumMapper : IObjectMapper
{
  public static TDestination 
      |\textbf{Map}|<TSource, TDestination>(TSource source)
  {
    var sourceEnumType = 
      ElementTypeHelper.GetEnumerationType(
        typeof(TSource));
    var destEnumType = 
      ElementTypeHelper.GetEnumerationType(
        typeof(TDestination));

    if (!Enum.IsDefined(sourceEnumType, source))
    {
      return (TDestination)Enum.ToObject(
        destEnumType, source);
    }

    if (!Enum.GetNames(destEnumType).Contains(
      source.ToString(),
      StringComparer.OrdinalIgnoreCase))
    {
      var underlyingSourceType = 
        Enum.GetUnderlyingType(|\slotbox{\boldmaroon{destEnumType}}|);
      var underlyingSourceValue = 
        System.Convert.ChangeType(
          source, underlyingSourceType);

      return (TDestination)Enum.ToObject(
        destEnumType, underlyingSourceValue);
    }

    return (TDestination)Enum.Parse(
      destEnumType, Enum.GetName(
        sourceEnumType, source), true);
  }

  private static readonly MethodInfo MapMethodInfo = 
    typeof(EnumToEnumMapper).GetAllMethods().First(
      _ => _.IsStatic);

  public bool |\textbf{IsMatch}|(TypePair |\boldmaroon{tbeqtxv}|)
  {
    var sourceEnumType = 
      ElementTypeHelper.GetEnumerationType(
        |\boldmaroon{tbeqtxv}|.SourceType);
    var destEnumType = 
      ElementTypeHelper.GetEnumerationType(
        |\boldmaroon{tbeqtxv}|.DestinationType);
    return sourceEnumType != null 
      && destEnumType != null;
  }
}
\end{minted}
\caption{}
\label{gnndestEnumTypeIncorrect}
\end{subfigure}

\end{tabular}
\end{minipage}
\caption{A C\# \sname{VarMisuse} example which is classified correctly as \scode{sourceEnumType} in the method \scode{Map} by the \sname{GGNN} model. Given the code and the target \scode{destEnumType}, our approach renames a local variable \scode{context} in \emph{another method}, making the model predict the wrong variable in the \scode{Map} method, thus introducing a real bug in the method \scode{Map}.}
\label{gnnDestinationType}
\end{figure*}